\pdfoutput=1

\documentclass[11pt]{article}

\usepackage[]{EMNLP2023}

\usepackage{times}
\usepackage{tabu}
\usepackage{latexsym}
\usepackage{stmaryrd}
\usepackage{amsmath}
\usepackage{multirow}
\usepackage{bbm}
\usepackage{graphicx}
\usepackage{amssymb}
\usepackage{booktabs}
\usepackage{algpseudocode}
\usepackage{algorithm}
\usepackage{graphicx}
\usepackage{caption}
\usepackage{subcaption}
\usepackage{tabularx}
\usepackage{colortbl}
\usepackage{xcolor}
\usepackage{tablefootnote}
\usepackage{enumitem}  
\usepackage{lipsum}
\usepackage{xspace}
\usepackage{bbding}
\usepackage{pifont}
\usepackage{arydshln}
\usepackage{array}
\usepackage{cleveref}
\newcommand{\PreserveBackslash}[1]{\let\temp=\\#1\let\\=\temp}
\newcolumntype{C}[1]{>{\PreserveBackslash\centering}p{#1}}
\newcolumntype{R}[1]{>{\PreserveBackslash\raggedleft}p{#1}}
\newcolumntype{L}[1]{>{\PreserveBackslash\raggedright}p{#1}}

\usepackage{hyperref}

\def\xHyphenate#1#2\wholeString {\if#1%
    \else\transform{#1}%
    \takeTheRest#2\ofTheString\fi}
\def\takeTheRest#1\ofTheString\fi
{\fi \xHyphenate#1\wholeString}
\def\transform#1{\url{#1}\hskip 0pt plus 1pt}

\usepackage[T1]{fontenc}

\usepackage{microtype}

\newcommand{\datasetname}{\texttt{ArabicMMLU}\xspace}

\definecolor{mygreen}{RGB}{217, 234, 211}
\definecolor{myred}{RGB}{244, 204, 204}

\usepackage{inconsolata}
\setlist{topsep=1pt,itemsep=1pt,partopsep=1pt,parsep=1pt}

\usepackage{arabtex}
\usepackage{utf8}

\title{ArabicMMLU:\\ Assessing Massive Multitask Language Understanding in Arabic}

\author{Fajri Koto$^{1}$ \, Haonan Li$^{1}$ \,   Sara Shatnawi$^{1}$ \,  Jad Doughman$^{1}$ \\  \textbf{Abdelrahman Boda Sadallah}$^{1}$ \,   \textbf{Aisha Alraeesi}$^{1}$ \,  \textbf{Khalid Almubarak}$^{2}$ \\   \textbf{Zaid Alyafeai}$^{3}$  \,  \textbf{Neha Sengupta}$^{4}$  \, \textbf{Shady Shehata}$^{1}$ \,  \textbf{Nizar Habash}$^{1,5}$ \\   \textbf{Preslav Nakov}$^{1}$  \,  \textbf{Timothy Baldwin}$^{1,6}$\\ 
$^{1}$Department of Natural Language Processing, MBZUAI \\
        $^{2}$Prince Sattam bin Abdulaziz University \, $^{3}$King Fahd University of Petroleum and Minerals \\
        $^{4}$Core42 \,  $^{5}$New York University  Abu Dhabi \, $^{6}$The University of Melbourne\\
	\texttt{\small \{fajri.koto,haonan.li,sara.shatnawi,jad.doughman,abdelrahman.sadallah,aisha.alraeesi\}@mbzuai.ac.ae 
	} 
}
\begin{document}
\setcode{utf8}

\maketitle
\begin{abstract}
    The focus of language model evaluation has transitioned towards reasoning and knowledge-intensive tasks, driven by advancements in pretraining large models. While state-of-the-art models are partially trained on large Arabic texts, evaluating their performance in Arabic remains challenging due to the limited availability of relevant datasets. To bridge this gap, we present \datasetname{}, the first multi-task language understanding benchmark for the Arabic language, sourced from school exams across diverse educational levels in different countries spanning North Africa, the Levant, and the Gulf regions. Our data comprises 40 tasks and 14,575 multiple-choice questions in Modern Standard Arabic (MSA) and is carefully constructed by collaborating with native speakers in the region. Our comprehensive evaluations of 35 models reveal substantial room for improvement, particularly among the best open-source models. Notably, BLOOMZ, mT0, LLaMA2, and Falcon struggle to achieve a score of 50\%, while even the top-performing Arabic-centric model only achieves a score of 62.3\%.\footnote{Data and code can be accessed at: \url{https://github.com/mbzuai-nlp/ArabicMMLU}}
\unskip\end{list}
\end{abstract}

\section{Introduction}

Although large language models (LLMs) such as GPT-3.5 \cite{ouyang2022training}, BLOOMZ \cite{muennighoff2022crosslingual}, and Jais \cite{sengupta2023jais} have been pretrained with substantial coverage of Modern Standard Arabic (MSA), their reasoning and knowledge assessments are primarily conducted using datasets translated from English to Arabic \cite{sengupta2023jais,huang2023acegpt}, which means there is limited capacity to evaluate content specific to Arabic. This reliance on translation systems not only demonstrates an Anglocentric approach \cite{ramesh-etal-2023-fairness,talat-etal-2022-reap} but also potentially introduces errors and biases. Given that Arabic is one of the most widely-spoken languages in the world, with a speaker population of over 400 million people \cite{shoufan-alameri-2015-natural,diab-etal-2017-nlp}, it is critically important that datasets be constructed for the language that are regionally- and culturally-localized.

\begin{figure}[t]
    \centering
    \includegraphics[width=\linewidth]{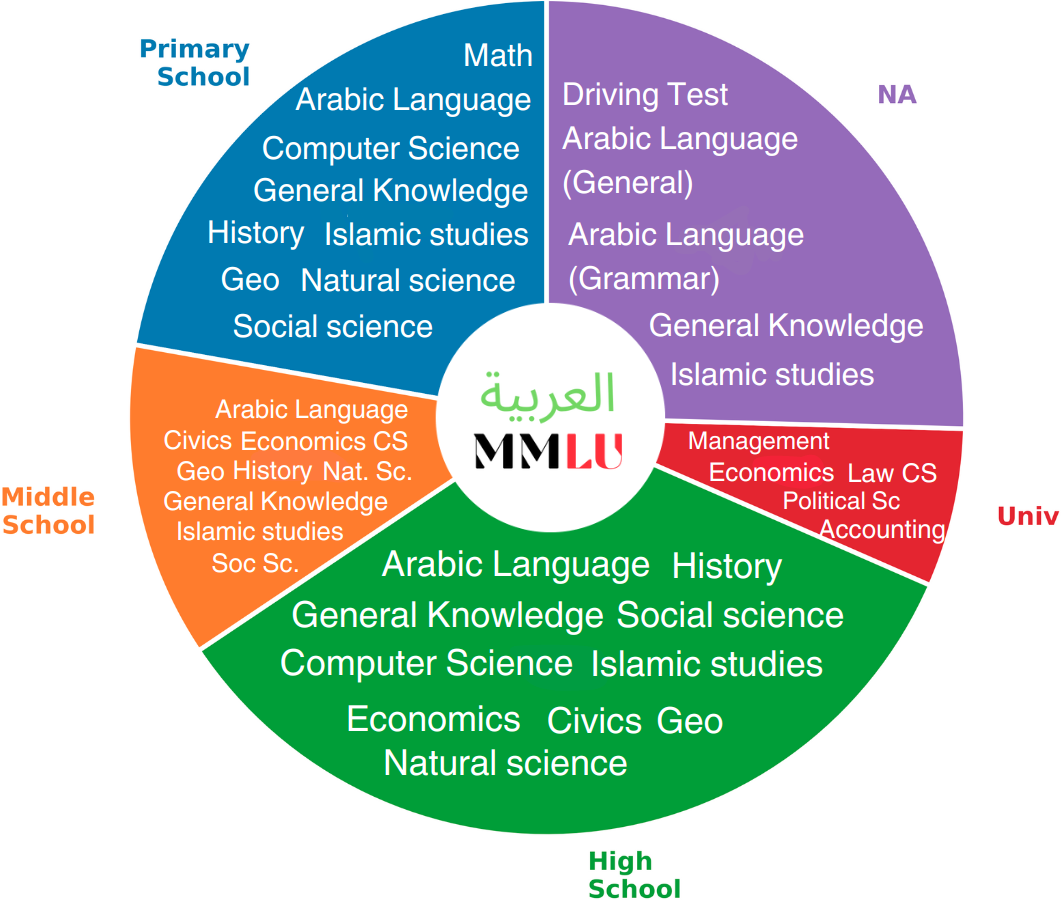} 
    \caption{Distribution of educational levels and corresponding subjects in \datasetname{}. ``NA'' denotes other levels.}
    \label{fig:overview}
\end{figure}

The evaluation of language models has increasingly shifted from linguistically-centric tasks, such as part-of-speech (POS) tagging and named entity recognition (NER), towards reasoning and knowledge evaluation. This shift is evidenced in evaluations of models like GPT-4 \cite{OpenAI2023GPT4TR}, LLaMA2 \cite{touvron2023llama2}, and LLM360 \cite{liu2023llm360} on various commonsense reasoning datasets \cite{zellers-etal-2019-hellaswag,huang-etal-2019-cosmos,koto-etal-2022-cloze,koto2024indoculture}, mathematical problems \cite{amini-etal-2019-mathqa,cobbe2021gsm8k}, coding challenges \cite{chen2021evaluating,austin2021program,yu2024codereval}, and school exams \cite{hendrycksmeasuring,li2023cmmlu,koto-etal-2023-large}. One notable dataset for knowledge evaluation is MMLU (Massive Multitask Language Understanding) \cite{hendrycksmeasuring}, which comprises multiple-choice questions across various subjects based on the US education system. In recent Arabic-centric LLMs like Jais \cite{sengupta2023jais} and AceGPT \cite{huang2023acegpt}, knowledge evaluation was carried out using MMLU translated from English to Arabic.

To comprehensively evaluate the reasoning and knowledge capabilities of Arabic LLMs in local Arabic-speaking contexts, we introduce \datasetname{}, styled around MMLU and sourced from school exams across Arabic-speaking countries spanning North Africa, the Levant, and the Gulf regions. \datasetname{} was constructed through collaboration with native Arabic speakers from Jordan, Egypt, UAE, Lebanon, and Saudi Arabia (KSA), ensuring rich local context, particularly in the subject areas of history, geography, law, civics education, and driving tests. \Cref{fig:overview} summarizes the distribution of education levels and corresponding subjects in \datasetname{}. The proportion of primary school, middle school, high school, and university level questions in \datasetname{} are 22.2\%, 12.2\%, 34\%, and 6.1\%, respectively, with the remaining questions categorized as ``NA''.

Our contributions can be summarized as follows:
\begin{itemize}
    \item We introduce the first Arabic MMLU-style dataset in Modern Standard Arabic (MSA), featuring 40 tasks covering various subject areas and educational levels across eight countries. Over 50\% of the questions in our dataset are tailored to Arabic-specific contexts.
    \item We evaluate 22 open-source multilingual models, 11 open-source Arabic-centric models, and 2 closed-source models. GPT-4 achieves the best performance, while the open-source models struggle to achieve scores above 60\%.
    \item We conduct a thorough analysis of the top-performing open-source models across various dimensions, encompassing: (1) individual subject areas, education levels, countries, and Arabic-specific topics; (2) few-shot inference performance; (3) model confidence; and (4) the influence of negation.
\end{itemize}


\section{Related Work}


\subsection{Language Models in Arabic} 

Early Arabic pretrained language models typically had less than 2 billion parameters and were primarily monolingual. These models can be classified into three main categories: encoder-only, decoder-only, and encoder--decoder models. The encoder-only models, such as AraBERT \cite{antoun-etal-2020-arabert}, CAMeLBERT \cite{inoue-etal-2021-interplay}, AraELECTRA \cite{antoun-etal-2021-araelectra}, and ARBERT~\&~MARBERT \cite{abdul-mageed-etal-2021-arbert}, are mainly from the BERT family. AraGPT2 \cite{antoun-etal-2021-aragpt2}, on the other hand, is a decoder-only model available in different sizes ranging from 135M to 1.4B parameters. Examples of encoder--decoder models include AraT5 \cite{nagoudi-etal-2022-arat5} and AraBART \cite{kamal-eddine-etal-2022-arabart}.

Jais~\cite{sengupta2023jais} and AceGPT \cite{huang2023acegpt} are two recent Arabic-centric decoder-only models with parameter sizes of up to 30B and 13B, respectively. Jais is pretrained on a corpus of 72 billion Arabic tokens, while AceGPT builds upon LLaMA2 and is enhanced with reinforcement learning from AI feedback \cite{rlaif} to localize the model to Arabic values and culture. Both models are bilingual (English and Arabic), and were fine-tuned on various instruction datasets.

Arabic is also present in multilingual models. This includes earlier models such as mBERT \cite{devlin2019bert} and XLM-R \cite{conneau-etal-2020-unsupervised}, and more recent LLMs such as BLOOMZ \cite{muennighoff2022crosslingual}, mT0  \cite{muennighoff2022crosslingual}, Falcon \cite{penedo2023refinedweb}, GPT-3.5 \cite{ouyang2022training}, and GPT-4 \cite{OpenAI2023GPT4TR}. In the original papers, only GPT-4 was evaluated in Arabic in terms of its reasoning and knowledge capabilities,  using the English--Arabic translated \texttt{MMLU} dataset, reporting an accuracy of 80\%.


\subsection{Arabic Benchmarks for Evaluating Language Models}

Arabic is included in various multilingual benchmarks for natural language understanding and generation, such as \texttt{XGLUE} \cite{liang-etal-2020-xglue}, \texttt{XTREME} \cite{hu2020xtreme}, \texttt{XTREME-R} \cite{ruder-etal-2021-xtreme} and \texttt{GEM} \cite{gehrmann-etal-2021-gem}. In recent years, several Arabic-centric benchmarks have been released, such as \texttt{Dolphin}~\cite{nagoudi-etal-2023-dolphin}, \texttt{OCRA}~\cite{elmadany2023orca}, and \texttt{LAraBench}~\cite{abdelali2024larabench}. Many tasks in these benchmarks involve classification, such as natural language inference \cite{conneau-etal-2018-xnli}, POS tagging \cite{darwish-etal-2017-arabic-pos}, named entity recognition \cite{pan-etal-2017-cross}, and summarization \cite{ladhak-etal-2020-wikilingua}. There are three notable question answering datasets: \texttt{TyDiQA} \cite{clark-etal-2020-tydi}, \texttt{Arabic-SQuAD} \cite{mozannar-etal-2019-neural}, and \texttt{MLQA} \cite{lewis-etal-2020-mlqa}. These datasets primarily focus on reading comprehension and question answering, unlike the \texttt{MMLU} dataset \cite{hendrycksmeasuring} which evaluates reasoning and knowledge in real-world settings, in the form of multiple-choice questions. Related, \texttt{EXAMs} \cite{hardalov-etal-2020-exams} is a dataset based on multilingual school exams, which contains a subset of about 500 Arabic questions.


\begin{figure}[t]
    \centering
    \includegraphics[width=\linewidth]{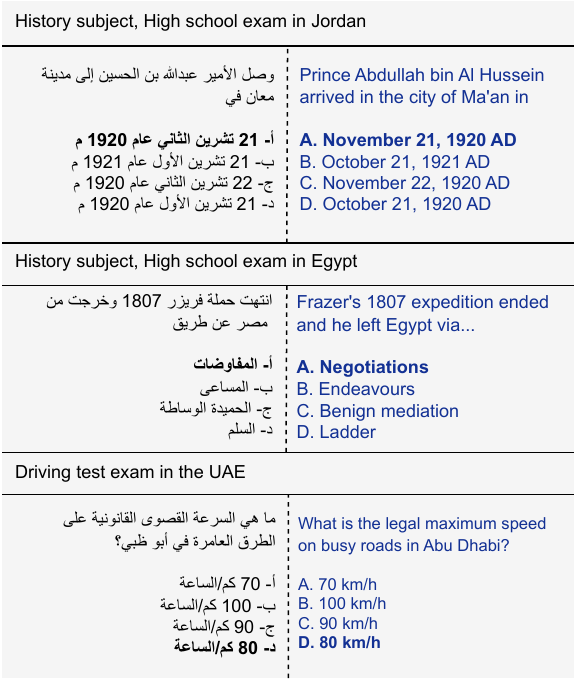} 
    \caption{Examples of two history questions and one driving test question from Jordan, Egypt, and UAE, respectively. \textbf{Left} is the original text and \textbf{right} is the English translation for illustrative purposes. The bold options are the correct answer keys.}
    \label{fig:ex1}
\end{figure}

\begin{figure}[ht]
    \centering
    \includegraphics[width=\linewidth]{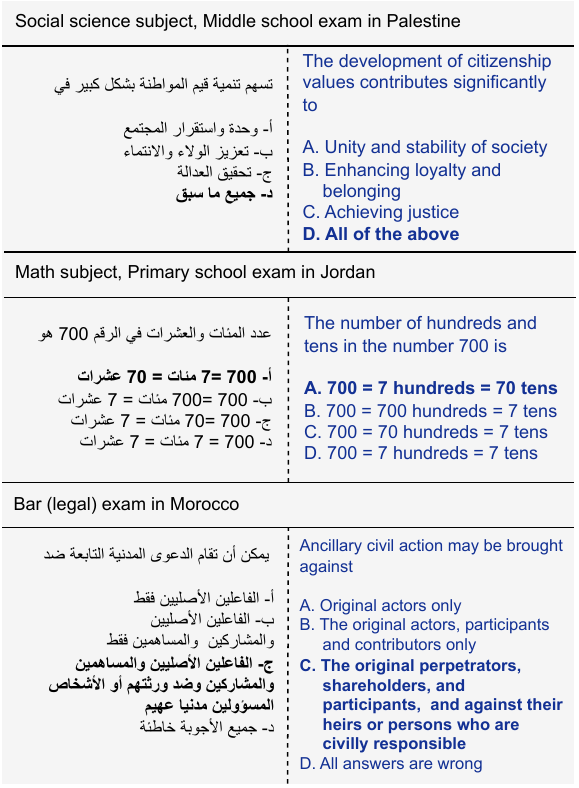} 
    \caption{Examples of social science, math, and bar exam questions from Palestine, Jordan, and Morocco, respectively. \textbf{Left} is the original text and \textbf{right} is the English translation for illustrative purposes. The bold options are the correct answer keys.}
    \label{fig:ex2}
\end{figure}

\section{ArabicMMLU}
\label{sec:data_construct}

\begin{table}[t]
    \centering
    \resizebox{\linewidth}{!}{
        \begin{tabular}{L{2.5cm}L{5.5cm}}
        \toprule
        \textbf{Group} & \textbf{Subjects} \\        
        \midrule
            STEM & Natural Science (P, M), Math (P), Physics (H), Biology (H), Computer Science (P, M, H, U)  \\ \midrule
            Social science & Social Science (P, M), Civics education (M, H), Geography (P, M, H), Economics (M, H, U), Political Science (U)  \\ \midrule
            Humanities & Islamic studies (P, M, H, U, NA), History (P, M, H), Accounting (U), Law (U),  Philosophy (H) \\ \midrule
            Language & Arabic Language (P, M, H), Arabic Language - General (NA), Arabic Language - Grammar (NA) \\ \midrule
            Other & Management (U), General Knowledge (P, M, NA), Driving Test (NA) \\
        \bottomrule
        \end{tabular}
    }
    \caption{Subject areas in \texttt{ArabicMMLU}. ``P'', ``M'', ``H'', ``U'', and , ``NA'' indicate that questions in the subject are available in primary school, middle school, high school, university and professional, and others, respectively.}
    \label{tab:group}
\end{table}

In the Middle East, the education system mostly follows the K12 system, consisting of six years of primary school, three years of middle school, and three years of high school.\footnote{\url{https://www.pwc.com/m1/en/industries/education/publications/understanding-middle-east-education.pdf}}$^,$\footnote{With the exception of the UAE, which follows a 4-4-4 structure for primary, middle, and high schools.} Many education systems in countries within the region, such as Egypt and KSA, prioritize Islamic studies alongside subjects like mathematics, natural science, social science, and geography.\footnote{\url{https://www.tabahfoundation.org/wp-content/uploads/2018/12/TabahFuturesInitiative-Islamic-Education_En.pdf}} In public schools, Arabic is commonly used for teaching and assessment, while in international schools, English is the predominant language of instruction for most subjects, following either the UK or USA curriculum. When designing \datasetname{}, we excluded questions in English and only included questions in Arabic.

\datasetname{} is an Arabic multiple-choice question-answering dataset comprising 40 tasks spanning a wide range of subjects and education levels. The questions are sourced from eight different countries in North Africa (Morocco and Egypt), the Levant (Jordan, Palestine, and Lebanon), and the Gulf (UAE, Kuwait, and KSA). Each question has 2--5 candidate answers, with one correct answer. \Cref{tab:group} provides details of the subjects in \datasetname{}. The subjects are drawn from different education levels (primary school, middle school, high school, university, and professional) and are categorized into STEM, social science, humanities, language, and other fields. 

\Cref{fig:ex1,fig:ex2} showcase various examples of \datasetname{} questions, with some focusing on history, driving tests, social science, and bar exams, all of which are pertinent to Arabic-specific norms and cultures. Notably, Arabic multiple-choice questions sometimes use Arabic-script characters
(\<أ>, 
\<ب>, 
\<ج>,
\<د>,
\<ه>)
rather than Latin-script characters (e.g.\ A, B, C, D, E). 
This differs from many other languages, where the answer options are strictly in Latin script (even if the local writing script is not Latin, as with Mandarin Chinese). In prior work \cite{hendrycksmeasuring,koto-etal-2023-large,li2023cmmlu}, answering these multiple-choice questions has relied on the probability of the alphabetic options. We experiment with both Arabic and Latin script outputs in \Cref{sec:exp}.

\subsection{Data Construction}

The data construction process involved a total of 10 Arabic native speakers from different countries: 6 internal workers (1 Jordanian, 1 Egyptian, 1 Lebanese, 1 from UAE, and 2 from KSA) and 4 external workers (3 Jordanian and 1 Egyptian).

During the first stage of data collection, the internal workers were tasked with collecting relevant sources for data collection. These sources were URLs containing the questions, which needed to be publicly available.

In the second stage, all workers were asked to manually scrape the data within a 2-month period. The task was to collect metadata, including the source (URL of the source document), country, subject, level, question, multiple-choice options, and the correct answer key. Each external worker was assigned to gather 2,000 questions, while internal workers were tasked with gathering 1,000--2,000 questions each. Our internal workers are Master's students and Research Assistants in Computer Science, while the external workers hold Bachelor's degrees. We ensured competitive compensation for the workers, exceeding the monthly average wage in each respective country.

During manual data scraping, workers were instructed to include only questions accompanied by an answer key, and to discard questions containing multi-modal information (e.g., images, videos, or tables). If a question had additional contextual information (e.g., a passage referenced by several questions), the context was required to be included with each corresponding question. 

\subsection{Quality Control}

While our workers are native speakers of Modern Standard Arabic with at least Bachelor's degrees, we maintain the quality of our dataset construction through meticulous steps. Firstly, we conducted a 1-hour workshop before the data collection stage to clarify the process. Secondly, we automatically filtered out repetitive questions and those without an answer key, reducing the initial set of over 15,000 questions to 14,575 unique questions. Finally, we assessed the accuracy of our data collection by having two native Arabic speakers annotate 100 randomly sampled questions. They were provided with all metadata, including the answer key, and tasked with verifying the correctness of each sample using any available resources (e.g., search engines). We found that 96\%  of the questions and answer keys match on average, while the remaining could have incorrect answer keys. This 96\% score is considered to represent the maximum score meaningfully achievable for \datasetname{}.

\subsection{Data Statistics}

\begin{table}[t]
    \centering
    \resizebox{\linewidth}{!}{
        \begin{tabular}{lrrr}
        \toprule
        \multirow{2}{*}{\textbf{Group}} & \multirow{2}{*}{\textbf{\# Questions}} & \multicolumn{2}{c}{\textbf{\# Chars}}\\
        \cmidrule{3-4}
        & & \textbf{Question} & \textbf{Answer} \\        
        \midrule
        Primary & 3239 & 43.6 & 30.4 \\
        Middle & 1775 & 58.3 & 54.6 \\
        High & 4963 & 76.7 & 66.0 \\
        Univ & 892 & 69.1 & 97.3 \\
        NA & 3706 & 311.4 & 52.7 \\ 
        \midrule
        STEM & 3220 & 60.0 & 49.4 \\
        Social Science & 3540 & 62.2 & 57.5 \\
        Humanities & 3655 & 57.1 & 60.2 \\
        Language & 1661 & 647.3 & 45.1 \\
        Other & 2499 & 57.7 & 59.1 \\
        \bottomrule
        \end{tabular}
    }
    \caption{Average question and answer length (in characters) for each education group and subject area.}
    \label{tab:stat}
\end{table}

\begin{table}[t]
    \centering
    \resizebox{\linewidth}{!}{
        \begin{tabular}{lrrrrrr}
        \toprule
        \textbf{Country} & \textbf{STEM} & \textbf{Social}  & \textbf{Hum.}  & \textbf{Lang.} & \textbf{Other}  & \textbf{TOTAL}\\        
        \midrule
            Jordan & 1086 & 2163 & 1579 & 362 & 863 & 6053 \\
            Egypt & 1012 & 581 & 335 & 324 & 254 & 2506 \\
            Palestine & 860 & 585 & 600 & 2 & -- & 2047 \\
            Morocco & -- & -- & 317 & -- & -- & 317 \\
            Lebanon & -- & -- & -- & -- & 239 & 239 \\
            UAE & -- & -- & 56 & -- & 128 & 184 \\
            Kuwait & -- & -- & -- & -- & 111 & 111 \\
            KSA & 67 & -- & -- & -- & 37 & 104 \\
            Other & 195 & 211 & 768 & 973 & 867 & 3014 \\
            \midrule
            TOTAL & 3220 & 3540 & 3655 & 1661 & 2499 & 14575 \\
        \bottomrule
        \end{tabular}
    }
    \caption{The distribution of \datasetname{} sources by country, categorized according to subject areas. ``Social'', ``Hum.'', and ``Lang.'' denote social science, humanities, and Arabic language, respectively.}
    \label{tab:stat_per_country}
\end{table}

\Cref{tab:stat} presents detailed statistics of \datasetname{}, categorized by education level and subject area. The distribution of questions across education levels varies, with primary school having the largest number, around 4.9K, followed by high school with 3.2K. Questions and candidate options are generally longer at the high school and university levels. Additionally, we observe that questions in the ``NA'' (other) category are four times longer (in characters) than those in school exams. This is expected since this category includes subjects like Arabic language (General) and Arabic language (Grammar), where questions typically involve lengthy paragraphs as context. For a detailed breakdown of questions for each subject in each education level, please refer to the Appendix (Table~\ref{tab:stat_per_subject}).

For subject areas, they are reasonably evenly distributed, particularly for STEM, social science, and the humanities, each consisting of roughly 3.2K to 3.5K questions. There are only minor differences in question length between these three subject areas. However, for the language category, the average question length (in characters) is 10 times longer than other categories.

\Cref{tab:stat_per_country} further shows the distribution of questions across the eight countries from which questions were collected, with Jordan, Egypt, and Palestine being the top three sources.  Various subjects within the social sciences, humanities, and other categories (such as driving tests) often include Arabic-specific content, representing 57.7\% of the dataset. While STEM questions are more aligned with the English \texttt{MMLU}, it is worth noting that differences in curriculum between North Africa, the Levant, the Gulf regions, and the USA may influence variations in assessment question design.

\section{{Experiments}}
\label{sec:exp}

\begin{figure}[t]
    \centering
    \includegraphics[width=0.85\linewidth]{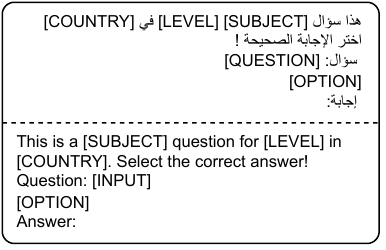} 
    \caption{Prompt templates in Arabic and English.}
    \label{fig:prompt}
\end{figure}

\begin{table*}[t]
    \centering
    \resizebox{0.95\linewidth}{!}{
        \begin{tabular}{lC{2cm}C{2cm}C{2cm}C{2cm}C{2cm}C{2cm}}
        \toprule
       \multirow{2}{*}{\textbf{Model (\#parameters)}}& \multirow{2}{*}{\textbf{STEM}} & \textbf{Social} & \multirow{2}{*}{\textbf{Humanities}} & \textbf{Arabic} &  \multirow{2}{*}{\textbf{Other}}  & \multirow{2}{*}{\textbf{Average}} \\
       & & \textbf{Science} & & \textbf{Language} & & \\
       \midrule
        Random & 29.5 & 28.9 & 28.6 & 25.8 & 32.3 & 29.0 \\
        \hdashline
        XGLM (1.7B) & 30.0 & 30.5 & 31.2 & 28.1 & 34.6 & 31.0 \\
        XGLM (2.9B) & 30.0 & 30.7 & 31.4 & 28.1 & 35.3 & 31.2 \\
        XGLM (4.5B) & 27.6 & 29.1 & 28.5 & 26.7 & 34.7 & 29.3 \\
        XGLM (7.5B) & 27.8 & 29.3 & 29.5 & 27.9 & 33.0 & 29.5 \\
        \hdashline
        BLOOMZ (560M) & 32.7 & 30.4 & 31.9 & 26.5 & 36.6 & 31.9 \\
        BLOOMZ (1.1B) & 30.4 & 26.5 & 30.1 & 25.1 & 28.1 & 28.4 \\
        BLOOMZ (1.7B) & 35.3 & 39.0 & 37.4 & 37.2 & 39.6 & 37.7 \\
        BLOOMZ (3B) & 40.4 & 44.5 & 43.8 & 40.9 & 48.5 & 43.7 \\
        BLOOMZ (7B) & 43.2 & 48.0 & 49.1 & 49.9 & 49.9 & 47.8 \\
        \hdashline
        mT0$_\text{small}$ (300M) & 31.1 & 30.5 & 29.4 & 29.4 & 33.2 & 30.7 \\
        mT0$_\text{base}$ (580M) & 30.2 & 30.9 & 31.5 & 28.2 & 34.4 & 31.2 \\
        mT0$_\text{large}$ (1.2B) & 31.1 & 31.7 & 31.6 & 29.7 & 35.7 & 32.0 \\
        mT0$_\text{xl}$ (3.7B) & 38.7 & 42.3 & 40.1 & 43.9 & 43.5 & 41.4 \\
        mT0$_\text{xxl}$ (13B ) & 42.7 & 45.4 & 43.4 & 46.0 & 46.0 & 44.5 \\
        \hdashline
        LLaMA2 (7B) & 33.7 & 32.8 & 33.5 & 28.4 & 36.7 & 33.4 \\
        LLaMA2-chat (7B) & 34.5 & 32.9 & 31.5 & 30.9 & 37.0 & 33.4 \\
        LLaMA2 (13B) & 32.9 & 35.0 & 37.8 & 35.8 & 39.3 & 36.1 \\
        LLaMA2-chat (13B) & 36.2 & 34.8 & 34.2 & 35.3 & 40.7 & 36.0 \\
        \hdashline
        Falcon (7B) & 29.8 & 29.9 & 31.5 & 29.0 & 35.1 & 31.1 \\
        Falcon-instruct (7B) & 28.4 & 29.5 & 27.3 & 21.3 & 29.1 & 27.7 \\
        Falcon (40B) & 34.9 & 33.8 & 36.2 & 30.1 & 37.4 & 34.8 \\
        Falcon-instruct (40B) & 33.8 & 30.9 & 33.9 & 28.9 & 36.2 & 33.0 \\
        \hdashline
        AraT5 (220M) & 29.9 & 30.3 & 33.0 & 28.4 & 32.0 & 31.0 \\
        AraT5v2 (220M) & 31.4 & 30.7 & 32.8 & 27.4 & 34.7 & 31.7 \\
        AraGPT2 (1.7B) & 33.0 & 31.5 & 35.8 & 29.8 & 37.4 & 33.7 \\ 
        \hdashline        
        AceGPT (7B) & 35.4 & 35.9 & 36.2 & 31.1 & 41.7 & 36.3 \\
        AceGPT-chat (7B) & 41.2 & 45.3 & 47.8 & 41.5 & 51.5 & 45.6 \\
        AceGPT (13B) & 42.7 & 45.5 & 48.3 & 42.4 & 50.7 & 46.1 \\
        AceGPT-chat (13B) & 47.3 & 52.8 & 53.9 & 50.5 & 58.5 & 52.6 \\
        \hdashline
        Jais (13B) & 30.3 & 31.4 & 33.6 & 28.1 & 36.3 & 32.2 \\
        Jais-chat (13B) & 50.5 & 56.1 & 59.3 & 39.9 & 61.7 & 54.8 \\
        Jais (30B) & 39.5 & 45.6 & 50.5 & 34.6 & 49.1 & 44.8 \\
        Jais-chat (30B) & \textbf{56.2} & \textbf{60.5} & \textbf{65.5} & \textbf{62.0} & \textbf{68.2} & \textbf{62.3} \\
        \midrule
        GPT-3.5 (175B) & 53.8 & 57.0 & 57.5 & 57.6 & 63.8 & 57.7 \\
        GPT-4 (NA) & \textbf{70.2} & \textbf{70.4} & \textbf{73.2} & \textbf{72.8} & \textbf{76.9} & \textbf{72.5} \\

        \bottomrule
        \end{tabular}
    }
    \caption{Zero-shot LLM performance (\% accuracy), combined across subject groups. ``Average'' means the average across all questions in \datasetname{}.}
    \label{tab:result}
\end{table*}

\subsection{Set-Up}
\label{sec:setup}

Our experiments focus on zero-shot and few-shot settings across 35 models. This includes 22 open-source multilingual models (XGLM~\cite{lin-etal-2022-shot},~BLOOMZ \cite{muennighoff2022crosslingual}, mT0~\cite{muennighoff2022crosslingual}, Falcon \cite{penedo2023refinedweb}, and LLaMA2~\cite{touvron2023llama2}, across various sizes), 11 open-source Arabic-centric models (AraT5~\cite{nagoudi-etal-2022-arat5}, AraGPT2~\cite{antoun-etal-2021-aragpt2}, AceGPT~\cite{huang2023acegpt} and Jais \cite{sengupta2023jais}, also across various sizes), and 2 closed-source models (GPT-3.5: \texttt{gpt-3.5-turbo} \cite{ouyang2022training} and GPT-4: \texttt{gpt-4-0613} \cite{OpenAI2023GPT4TR}).

\begin{figure}[t]
    \centering
    \includegraphics[width=\linewidth]{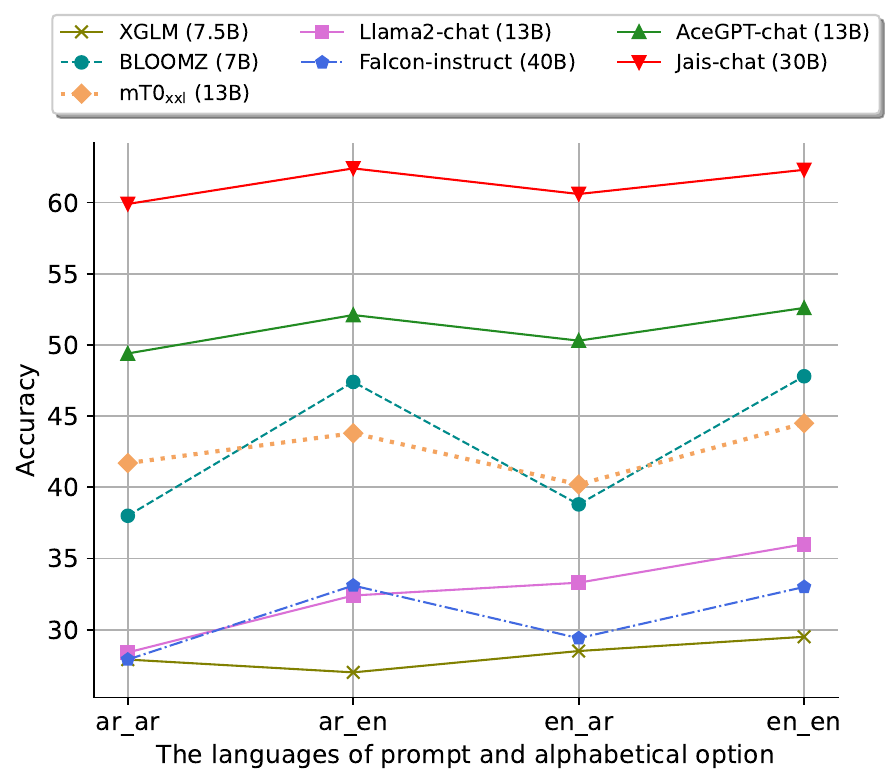} 
    \caption{LLM peformance with different prompt settings. \texttt{ar\_en} means that the prompt template is in Arabic and the alphabetic option is in English (the Latin script).}
    \label{fig:benchmark}
\end{figure}

We initially conducted experiments with four settings: (1) Arabic prompt and Arabic alphabetic output, (2) Arabic prompt and English (i.e.\ Latin script) alphabetic output, (3) English prompt and Arabic alphabetic output, and (4) English prompt and English alphabetic output. \Cref{fig:prompt} illustrates the Arabic and English prompts. The placeholders [SUBJECT], [LEVEL], and [COUNTRY] are replaced with the corresponding Arabic and English words, while the placeholders [INPUT] and [OPTION] are in Arabic. The choice of the alphabetic output (English vs.\ Arabic) is adjusted in [OPTION]. See Appendix~\ref{app:examples} (Figure~\ref{fig:prompt_example}) for examples of the full input in both English and Arabic.

Following previous studies \cite{koto-etal-2023-large,li2023cmmlu}, for open-source models, we determine the answer based on the highest probability among all possible options. In the case of English alphabetic output, we measure the probability of the first generated token being A, B, C, D, or E. For Arabic, we measure the probability of the first generated token being
\<أ>, 
\<ب>, 
\<ج>,
\<د>, or
\<ه>.
For closed-source models, we determine the answer based on the first token generated in the text using a regular expression. If there is no match, we assign a random answer.




\subsection{Results}
\label{sec:main-results}

To evaluate the influence of prompt language, we initially benchmarked the open-source models using all four prompt settings (\Cref{sec:setup}), as depicted in \Cref{fig:benchmark}.  We observe that the optimal configuration across all models is to use an English prompt and English alphabetic output. Predictably, the Arabic-specific LLMs --- Jais-chat (30B) and AceGPT-chat (13B) --- demonstrate the greatest robustness when employing Arabic alphabetic output. Please refer to Appendix for complete results of all prompt settings across the open-source models. For the remaining experiments, we will report based on the setting of English prompt and English alphabetic output.

\paragraph{Results across all models}
\Cref{tab:result} shows the full results of all models, grouped by subject area.
As expected, the Arabic-centric model Jais-chat (30B) emerges as the top-performing open-source model, boasting an average score of 62.3\%, surpassing GPT-3.5 by 4.6 points. Compared to AceGPT-chat (13B), both Jais-chat models (13B and 30B) exhibit substantially higher accuracy in areas including STEM, Social Science, Humanities, and Others. For multilingual models such as BLOOMZ (7B) and mT0 (13B), their performance lags behind Jais, with a disparity of more than 14 points. XGLM, LLaMA2, and Falcon perform at a level close to random, suggesting their limited proficiency in Arabic. GPT-4 achieves the highest accuracy, with a score of 72.5\%, surpassing Jais-chat (30B) by 10 points. It is noteworthy that in the GPT-4 technical report \cite{OpenAI2023GPT4TR}, the accuracy of the English-Arabic translated MMLU dataset is reported as 80\%, which is 8 points higher than our \datasetname{} results. One possible explanation for this difference is that our \datasetname{} presents a greater challenge due to its inclusion of a higher proportion of Arabic-specific content.

Furthermore, we notice a trend of increasing accuracy with larger models, with the exception of XGLM. For example, BLOOMZ (7B) achieves an accuracy 15.9 points higher than BLOOMZ (560M), while mT0 (13B) shows a 13.8-point increase compared to mT0 (300M). This trend is also evident in AceGPT and Jais, although it is less pronounced in LLaMA2 and Falcon, which are English-centric models.

\begin{figure}[t]
    \centering
    \includegraphics[width=\linewidth]{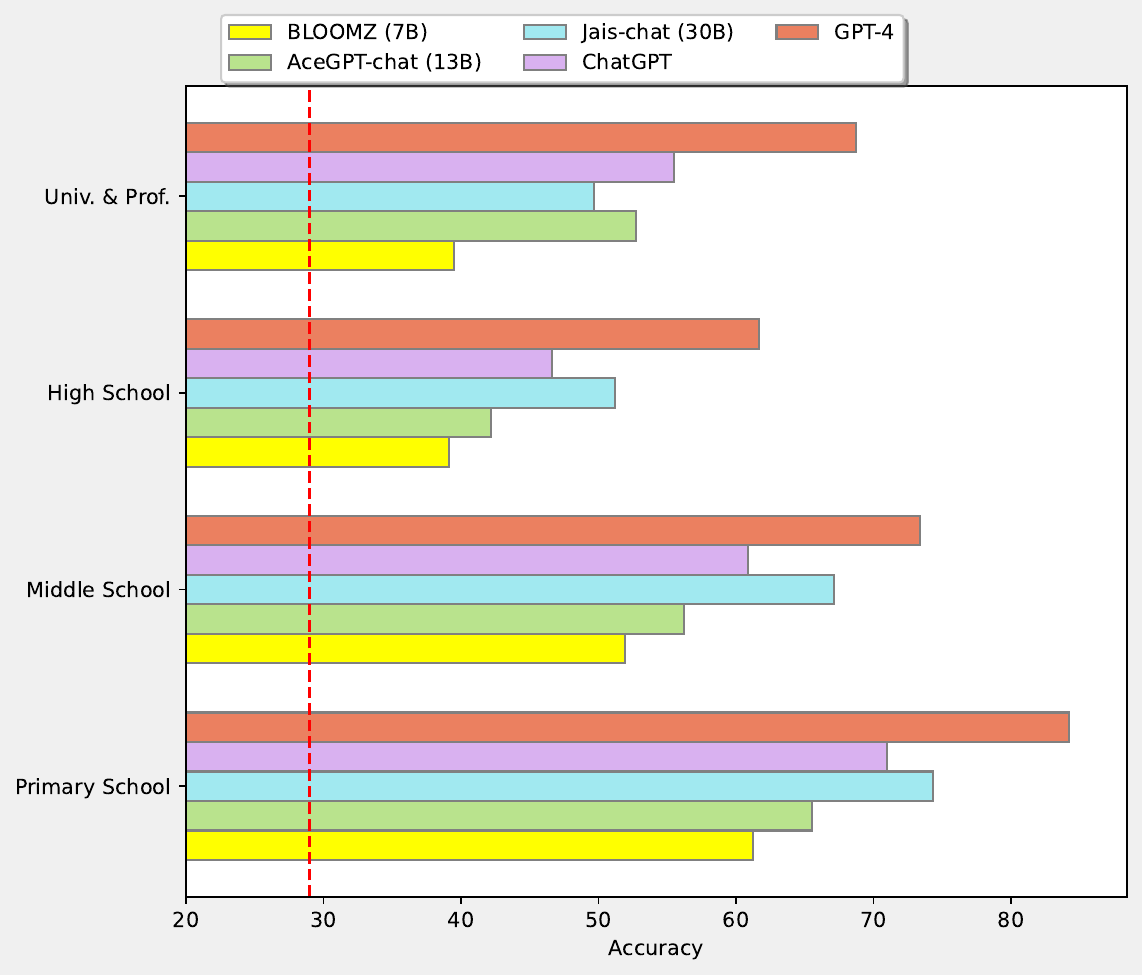} 
    \caption{LLM performance across different education levels.}
    \label{fig:education}
\end{figure}

\paragraph{Results across education levels}
\Cref{fig:education} depicts the average scores of the top-performing models (BLOOMZ, AceGPT-chat, Jais-chat, GPT-3.5, and GPT-4) across different education levels. We observe that \datasetname{} questions are more challenging at the high school level compared to the primary and middle school levels. Specifically, for high school questions, GPT-4 achieves a score of only 61.7\%, while Jais-chat scores 51.2\%. Interestingly, we notice that the model accuracy at the university level is higher than for high school. This could be attributed to the relatively small portion (i.e., 6\%) of university-level questions in \datasetname{}, which potentially skews the results.

\begin{table}[t]
    \centering
    \resizebox{\linewidth}{!}{
        \begin{tabular}{lrccc}
        \toprule
        \textbf{Country} & \textbf{\# Q.} & \textbf{BLOOMZ} & \textbf{{AceGPT}} & \textbf{Jais}\\    
        \midrule
        UAE & 128 & 29.7 & 46.9 & \textbf{48.4} \\
        Egypt & 830 & 45.0 & 48.0 & \textbf{55.6} \\
        Lebanon & 239 & 55.6 & 62.8 & \textbf{69.5} \\
        Jordan & 2532 & 45.6 & 51.5 & \textbf{59.8} \\
        Kuwait & 111 & 44.1 & 53.2 & \textbf{58.6} \\
        KSA & 37 & 32.4 & 54.1 & \textbf{56.8} \\
        Palestine & 152 & 42.1 & 52.6 & \textbf{63.8} \\
        Morocco & 314 & 25.9 & \textbf{52.7} & 33.1 \\       
        \bottomrule
        \end{tabular}
    }
    \caption{Average performance on subjects with Arabic-specific context, grouped by countries. Here we use BLOOMZ (7B), AceGPT-chat (13B), and Jais-chat (30B).}
    \label{tab:result_country}
    \vspace{-0.2cm}
\end{table}

\paragraph{Results by country} We present the performance of open-source models on selected subjects that potentially contain Arabic-specific contexts. These subjects include history, geography, civics, political science, law, and driving tests, grouped by country in \Cref{tab:result_country}. We observe that BLOOMZ performs less well on questions sourced from the UAE and Morocco compared to other countries, while Jais performs best overall except in questions sourced from Morocco. 

\subsection{Analysis}
We focus our more detailed analysis in this section solely on the best open-source models, namely BLOOMZ, AceGPT, and Jais, providing researchers and the community with insights to better understand these models and opportunities for future improvements.

\paragraph{Few-shot performance} 

\begin{figure}[t]
    \centering
    \includegraphics[width=\linewidth]{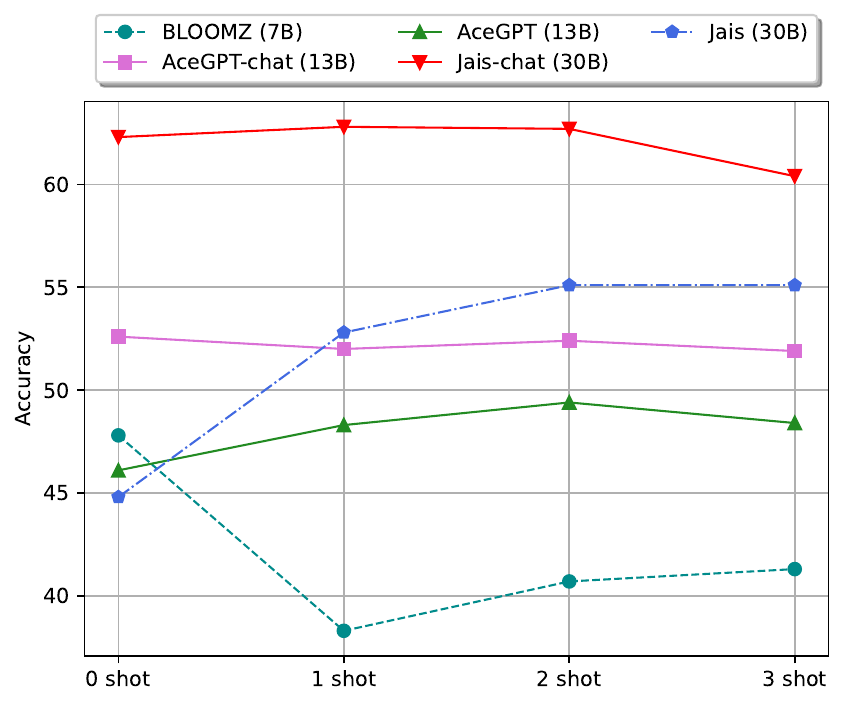} 
    \caption{Few-shot performance (\% accuracy) of LLMs averaged across all questions.}
    \label{fig:fewshot}
\end{figure}

While all the results in \Cref{sec:main-results} were based on zero-shot learning, we observe in \Cref{fig:fewshot} that when we move to few-shot learning, results for base models improve but those for instruction-tuned models deteriorate. Specifically, AceGPT and Jais show an improvement of 2--10 points when using few-shot learning, but the results for BLOOMZ and Jais-chat drop. These findings are consistent with prior research over \texttt{IndoMMLU}~\cite{koto-etal-2023-large} and \texttt{CMMLU}~\cite{li2023cmmlu}.

\paragraph{Model confidence}

\begin{figure}[t]
    \centering
    \includegraphics[width=\linewidth]{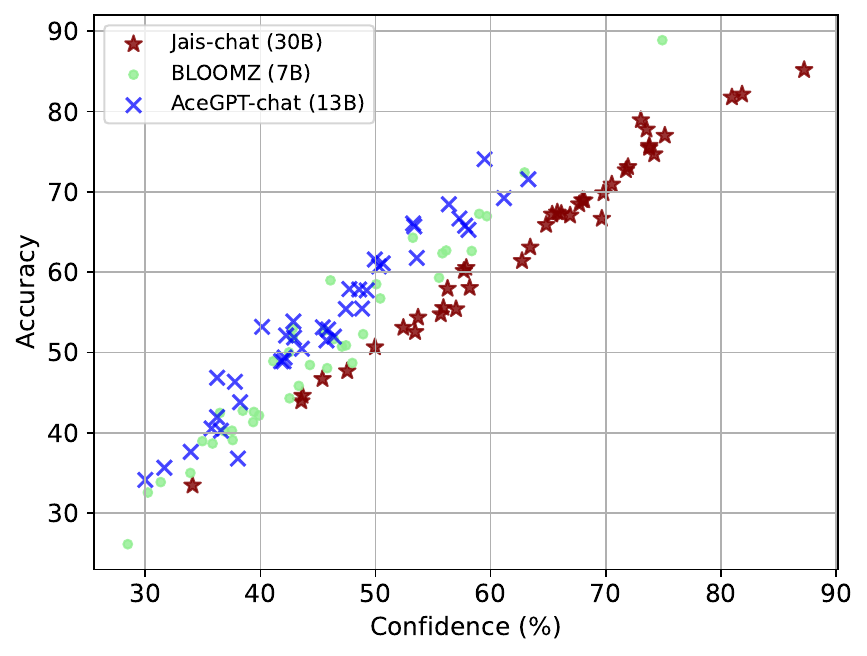} 
    \caption{Zero-shot calibration of BLOOMZ, AceGPT-chat, and Jais-chat across 40 tasks. Confidence (\%) denotes the average probability scores in percentage.}
    \label{fig:conf1}
     \vspace{-0cm}
\end{figure}

We analyze whether BLOOMZ, AceGPT, and Jais are well-calibrated in answering \datasetname{} questions by comparing the probability of the correct answers with the actual accuracy for each task (i.e., subject and level combination). The answer probability is obtained through softmax normalization across the available candidate answers. In \Cref{fig:conf1}, we observe that the three open-source models are well calibrated with correlation scores $r > 0.9$.

\begin{figure}[t]
    \centering
    \includegraphics[width=\linewidth]{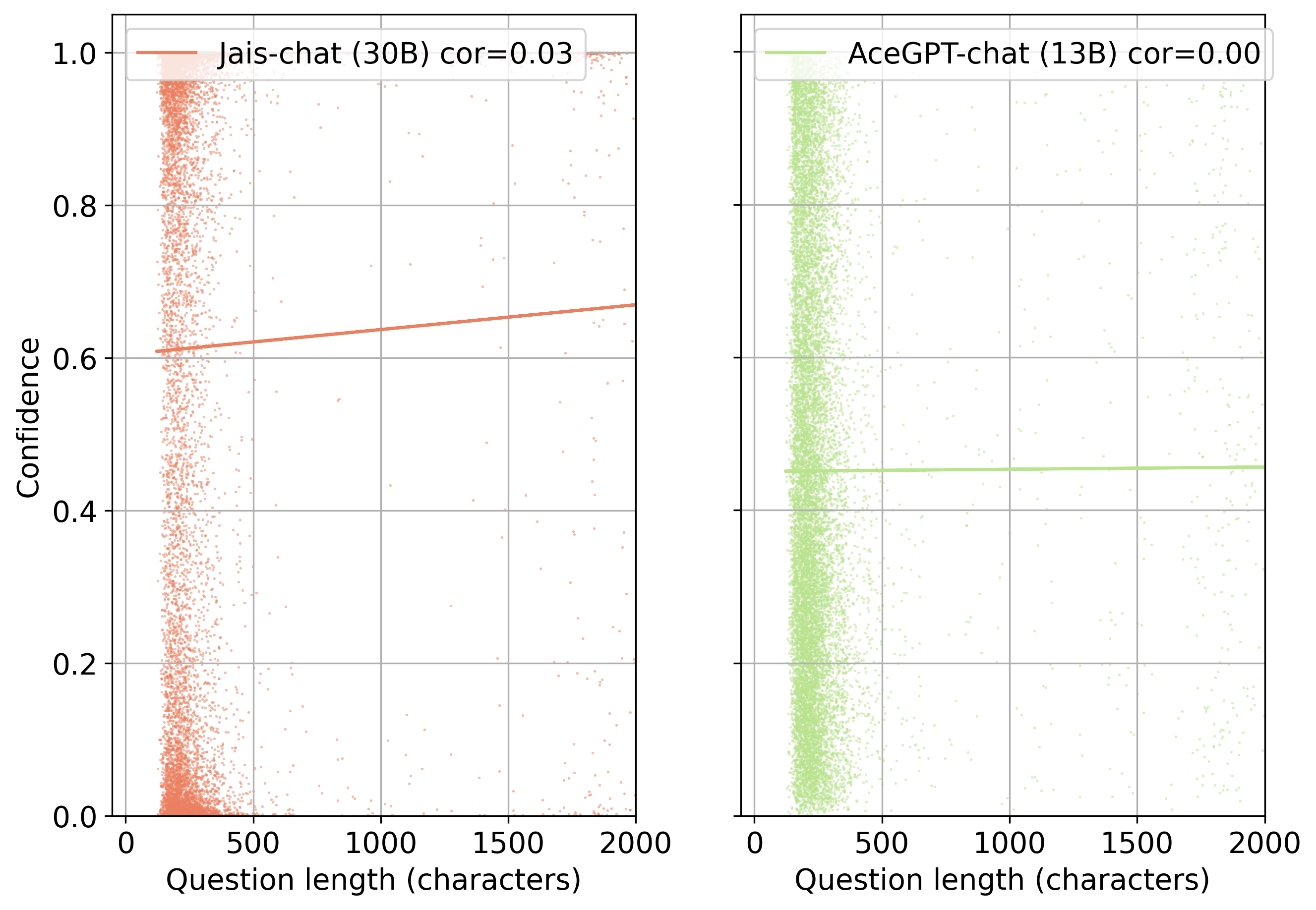} 
    \caption{Correlation between model confidence and question length.}
    \label{fig:conf2}
\end{figure}

Additionally, we investigate the correlation between model confidence and question length in \Cref{fig:conf2}. We find no correlation between the length of the questions and the model confidence for either Jais or AceGPT.

\paragraph{Impact of negation}

Despite negation being an absolutely foundational linguistic phenomenon, LLMs have been shown to be worryingly insensitive to its effects in English \citep{DBLP:conf/acl/KassnerS20,DBLP:conf/naacl/HosseiniRBHSC21,truong-etal-2023-language}.
We thus perform an analysis of LLM performance over questions in \datasetname{} with and without negation to determine whether this observation ports across to Arabic.
We utilize specific negation phrases to identify questions containing negations in Arabic. These include:
\<لا> (no),
\<ليس> (is not),
\<ليست> (is not),
\<لم> (did not), 
\<من غير> (without),
\<باستثناء> (excluding), and
\<دون> (without).
To prevent ambiguity, the term \<ما> is omitted, as it can also mean ``what''. After applying this filtering, we obtain 816 questions. We randomly inspected 100 random samples and found the detection accuracy for negation to exceed 95\%.


\Cref{tab:negation} presents the accuracy of the LLMs in answering questions with and without negation in the top three subjects containing negation (Geography, Biology, and Economics). Overall, negated questions generally exhibit slightly lower accuracy, particularly in Biology and Economics. However, for Geography, the models actually achieve higher accuracy.


\begin{table}[t]
    \centering
    \resizebox{\linewidth}{!}{
        \begin{tabular}{lcc}
        \toprule
            \textbf{Model} & \textbf{W/o negation} & \textbf{W/ negation} \\
            \midrule
            \multicolumn{3}{l}{\textit{Geography} (8.0\%)}  \\ 
            BLOOMZ (7B)& \textbf{42.8} & 42.2 \\
            AceGPT-chat (13B)  & 48.7 & \textbf{53.2} \\
            Jais-chat (30B)  & \textbf{56.9} & 48.6 \\
            \midrule
            \multicolumn{3}{l}{\textit{Biology} (6.7\%)}  \\
            BLOOMZ (7B) & \textbf{35.2} & 31.6 \\
            AceGPT-chat (13B) & \textbf{37.7} & 35.8 \\
            Jais-chat (30B) & \textbf{47.0} & 43.2 \\
            \midrule
            \multicolumn{3}{l}{\textit{Economics} (13.3\%)}  \\
            BLOOMZ (7B) & \textbf{52.5} & 37.7 \\
            AceGPT-chat (13B) & 50.1 & \textbf{51.9} \\
            Jais-chat (30B) & \textbf{60.4} & 54.5 \\
        \bottomrule
        \end{tabular}
    }
    \caption{Model accuracy in answering questions with and without negation in Geography, Biology, and Economics. The number following the subject name indicates the proportion of negated questions.}
    \label{tab:negation}
\end{table}

\subsection{Discussion}

Our experiments show that open-source LLMs perform poorly on \datasetname{} questions, particularly multilingual models. Furthermore, the Arabic-centric LLMs still struggle to capture Arabic cultural knowledge across all education levels. This highlights a significant need for improvement in Arabic LLMs.  In contrast, GPT-4 demonstrates remarkable performance across all tasks, surpassing all other models. However, it remains unclear whether the success of GPT-4 results from scaling up the dataset and model size or simply from memorization (given that all questions were taken from public sources).

\section{Conclusion and Future Work}

We introduce \datasetname{}, the first large-scale multi-task language understanding dataset designed to evaluate real-world knowledge in Arabic. Through experiments with over 14K multiple-choice questions spanning various subjects and education levels, we observed that Arabic-centric LLMs outperform multilingual LLMs, albeit with lower accuracy than GPT-4. We envision \datasetname{} as a valuable resource for tracking the real-world knowledge and reasoning capabilities of future Arabic LLMs. For future work, \datasetname{} can be extended to include short-answer or essay questions, different modalities (i.e., images, audio, video), larger region coverage, and more questions in professional settings. This will enhance the evaluation to better reflect real-world scenarios.


\section*{Limitations}

Although we believe our benchmark will significantly contribute to the advancement of Arabic LLMs, it is important to acknowledge limitations that need to be addressed in future work. We outline these limitations as follows:

\paragraph{Limited diversity} \datasetname{} does not represent all Arabic countries equally. For example, we have collected over 6K multiple-choice questions from Jordan, while other countries are represented with only 100 questions or, in some cases, not at all.  This is largely due to the availability of publicly-accessible exams in each country; some countries have digitized their exams, but not others. Additionally, our search for relevant Arabic content across the internet was not exhaustive. 

\paragraph{Dialectical Arabic is excluded} 
The dataset primarily focuses on Modern Standard Arabic (MSA). However, multilingual and Arabic LLMs are often exposed to both MSA and dialectical Arabic. 



\paragraph{Text-based questions only} 
\datasetname{} is focused solely on text-based assessment, and the exploration of multimodal questions is left for future work.


\section*{Ethical Considerations}

It is important to emphasize that our experimental results do not provide conclusive answers regarding the performance of LLMs in Arabic. This issue becomes even more vexing when discussing the GPT-4 results, which outperformed all models, due to a lack of sufficient information about its training regimen. As such, we cannot assert that the model's pretraining data is free from contamination.


\section*{Acknowledgements}

We extend our gratitude to all collaborators from Jordan, Egypt, Lebanon, UAE, and Saudi Arabia who participated in the data collection process. We also acknowledge the contributions of Samta Kamboj, Sarah Al Barri, and Onkar Pandit from Core42, who assisted in collecting the Arabic Language question dataset.

\bibliography{custom,anthology}

\begin{thebibliography}{54}
\expandafter\ifx\csname natexlab\endcsname\relax\def\natexlab#1{#1}\fi

\bibitem[{Abdelali et~al.(2024)Abdelali, Mubarak, Chowdhury, Hasanain, Mousi, Boughorbel, Kheir, Izham, Dalvi, Hawasly, Nazar, Elshahawy, Ali, Durrani, Milic-Frayling, and Alam}]{abdelali2024larabench}
Ahmed Abdelali, Hamdy Mubarak, Shammur~Absar Chowdhury, Maram Hasanain, Basel Mousi, Sabri Boughorbel, Yassine~El Kheir, Daniel Izham, Fahim Dalvi, Majd Hawasly, Nizi Nazar, Yousseif Elshahawy, Ahmed Ali, Nadir Durrani, Natasa Milic-Frayling, and Firoj Alam. 2024.
\newblock \href {http://arxiv.org/abs/2305.14982} {{LA}ra{B}ench: Benchmarking {A}rabic {AI} with large language models}.
\newblock arXiv preprint arXiv:2305.14982.

\bibitem[{Abdul-Mageed et~al.(2021)Abdul-Mageed, Elmadany, and Nagoudi}]{abdul-mageed-etal-2021-arbert}
Muhammad Abdul-Mageed, AbdelRahim Elmadany, and El~Moatez~Billah Nagoudi. 2021.
\newblock \href {https://doi.org/10.18653/v1/2021.acl-long.551} {{ARBERT} {\&} {MARBERT}: Deep bidirectional transformers for {A}rabic}.
\newblock In \emph{Proceedings of the 59th Annual Meeting of the Association for Computational Linguistics and the 11th International Joint Conference on Natural Language Processing (Volume 1: Long Papers)}, pages 7088--7105, Online. Association for Computational Linguistics.

\bibitem[{Amini et~al.(2019)Amini, Gabriel, Lin, Koncel-Kedziorski, Choi, and Hajishirzi}]{amini-etal-2019-mathqa}
Aida Amini, Saadia Gabriel, Shanchuan Lin, Rik Koncel-Kedziorski, Yejin Choi, and Hannaneh Hajishirzi. 2019.
\newblock \href {https://doi.org/10.18653/v1/N19-1245} {{M}ath{QA}: Towards interpretable math word problem solving with operation-based formalisms}.
\newblock In \emph{Proceedings of the 2019 Conference of the North {A}merican Chapter of the Association for Computational Linguistics: Human Language Technologies, Volume 1 (Long and Short Papers)}, pages 2357--2367, Minneapolis, Minnesota. Association for Computational Linguistics.

\bibitem[{Antoun et~al.(2020)Antoun, Baly, and Hajj}]{antoun-etal-2020-arabert}
Wissam Antoun, Fady Baly, and Hazem Hajj. 2020.
\newblock \href {https://aclanthology.org/2020.osact-1.2} {{A}ra{BERT}: Transformer-based model for {A}rabic language understanding}.
\newblock In \emph{Proceedings of the 4th Workshop on Open-Source Arabic Corpora and Processing Tools, with a Shared Task on Offensive Language Detection}, pages 9--15, Marseille, France. European Language Resource Association.

\bibitem[{Antoun et~al.(2021{\natexlab{a}})Antoun, Baly, and Hajj}]{antoun-etal-2021-araelectra}
Wissam Antoun, Fady Baly, and Hazem Hajj. 2021{\natexlab{a}}.
\newblock \href {https://aclanthology.org/2021.wanlp-1.20} {{A}ra{ELECTRA}: Pre-training text discriminators for {A}rabic language understanding}.
\newblock In \emph{Proceedings of the Sixth Arabic Natural Language Processing Workshop}, pages 191--195, Kyiv, Ukraine (Virtual). Association for Computational Linguistics.

\bibitem[{Antoun et~al.(2021{\natexlab{b}})Antoun, Baly, and Hajj}]{antoun-etal-2021-aragpt2}
Wissam Antoun, Fady Baly, and Hazem Hajj. 2021{\natexlab{b}}.
\newblock \href {https://aclanthology.org/2021.wanlp-1.21} {{A}ra{GPT}2: Pre-trained transformer for {A}rabic language generation}.
\newblock In \emph{Proceedings of the Sixth Arabic Natural Language Processing Workshop}, pages 196--207, Kyiv, Ukraine (Virtual). Association for Computational Linguistics.

\bibitem[{Austin et~al.(2021)Austin, Odena, Nye, Bosma, Michalewski, Dohan, Jiang, Cai, Terry, Le et~al.}]{austin2021program}
Jacob Austin, Augustus Odena, Maxwell Nye, Maarten Bosma, Henryk Michalewski, David Dohan, Ellen Jiang, Carrie Cai, Michael Terry, Quoc Le, et~al. 2021.
\newblock Program synthesis with large language models.
\newblock \emph{arXiv preprint arXiv:2108.07732}.

\bibitem[{Chen et~al.(2021)Chen, Tworek, Jun, Yuan, Pinto, Kaplan, Edwards, Burda, Joseph, Brockman et~al.}]{chen2021evaluating}
Mark Chen, Jerry Tworek, Heewoo Jun, Qiming Yuan, Henrique Ponde de~Oliveira Pinto, Jared Kaplan, Harri Edwards, Yuri Burda, Nicholas Joseph, Greg Brockman, et~al. 2021.
\newblock Evaluating large language models trained on code.
\newblock \emph{arXiv preprint arXiv:2107.03374}.

\bibitem[{Clark et~al.(2020)Clark, Choi, Collins, Garrette, Kwiatkowski, Nikolaev, and Palomaki}]{clark-etal-2020-tydi}
Jonathan~H. Clark, Eunsol Choi, Michael Collins, Dan Garrette, Tom Kwiatkowski, Vitaly Nikolaev, and Jennimaria Palomaki. 2020.
\newblock \href {https://doi.org/10.1162/tacl_a_00317} {{T}y{D}i {QA}: A benchmark for information-seeking question answering in typologically diverse languages}.
\newblock \emph{Transactions of the Association for Computational Linguistics}, 8:454--470.

\bibitem[{Cobbe et~al.(2021)Cobbe, Kosaraju, Bavarian, Chen, Jun, Kaiser, Plappert, Tworek, Hilton, Nakano, Hesse, and Schulman}]{cobbe2021gsm8k}
Karl Cobbe, Vineet Kosaraju, Mohammad Bavarian, Mark Chen, Heewoo Jun, Lukasz Kaiser, Matthias Plappert, Jerry Tworek, Jacob Hilton, Reiichiro Nakano, Christopher Hesse, and John Schulman. 2021.
\newblock Training verifiers to solve math word problems.
\newblock \emph{arXiv preprint arXiv:2110.14168}.

\bibitem[{Conneau et~al.(2020)Conneau, Khandelwal, Goyal, Chaudhary, Wenzek, Guzm{\'a}n, Grave, Ott, Zettlemoyer, and Stoyanov}]{conneau-etal-2020-unsupervised}
Alexis Conneau, Kartikay Khandelwal, Naman Goyal, Vishrav Chaudhary, Guillaume Wenzek, Francisco Guzm{\'a}n, Edouard Grave, Myle Ott, Luke Zettlemoyer, and Veselin Stoyanov. 2020.
\newblock \href {https://doi.org/10.18653/v1/2020.acl-main.747} {Unsupervised cross-lingual representation learning at scale}.
\newblock In \emph{Proceedings of the 58th Annual Meeting of the Association for Computational Linguistics}, pages 8440--8451, Online. Association for Computational Linguistics.

\bibitem[{Conneau et~al.(2018)Conneau, Rinott, Lample, Williams, Bowman, Schwenk, and Stoyanov}]{conneau-etal-2018-xnli}
Alexis Conneau, Ruty Rinott, Guillaume Lample, Adina Williams, Samuel Bowman, Holger Schwenk, and Veselin Stoyanov. 2018.
\newblock \href {https://doi.org/10.18653/v1/D18-1269} {{XNLI}: Evaluating cross-lingual sentence representations}.
\newblock In \emph{Proceedings of the 2018 Conference on Empirical Methods in Natural Language Processing}, pages 2475--2485, Brussels, Belgium. Association for Computational Linguistics.

\bibitem[{Darwish et~al.(2017)Darwish, Mubarak, Abdelali, and Eldesouki}]{darwish-etal-2017-arabic-pos}
Kareem Darwish, Hamdy Mubarak, Ahmed Abdelali, and Mohamed Eldesouki. 2017.
\newblock \href {https://doi.org/10.18653/v1/W17-1316} {{A}rabic {POS} tagging: Don{'}t abandon feature engineering just yet}.
\newblock In \emph{Proceedings of the Third {A}rabic Natural Language Processing Workshop}, pages 130--137, Valencia, Spain. Association for Computational Linguistics.

\bibitem[{Devlin et~al.(2019)Devlin, Chang, Lee, and Toutanova}]{devlin2019bert}
Jacob Devlin, Ming-Wei Chang, Kenton Lee, and Kristina Toutanova. 2019.
\newblock {BERT}: Pre-training of deep bidirectional transformers for language understanding.
\newblock In \emph{Proceedings of the 2019 Conference of the North American Chapter of the Association for Computational Linguistics: Human Language Technologies, Volume 1 (Long and Short Papers)}, pages 4171--4186.

\bibitem[{Diab et~al.(2017)Diab, Habash, and Zitouni}]{diab-etal-2017-nlp}
Mona Diab, Nizar Habash, and Imed Zitouni. 2017.
\newblock \href {https://aclanthology.org/2017.tal-3.2} {{NLP} for {A}rabic and related languages}.
\newblock \emph{Traitement Automatique des Langues}, 58(3):9--13.

\bibitem[{Elmadany et~al.(2023)Elmadany, Nagoudi, and Abdul-Mageed}]{elmadany2023orca}
AbdelRahim Elmadany, El~Moatez~Billah Nagoudi, and Muhammad Abdul-Mageed. 2023.
\newblock \href {http://arxiv.org/abs/2212.10758} {{ORCA}: A challenging benchmark for arabic language understanding}.
\newblock arXiv preprint arXiv:2212.10758.

\bibitem[{Gehrmann et~al.(2021)Gehrmann, Adewumi, Aggarwal, Ammanamanchi, Aremu, Bosselut, Chandu, Clinciu, Das, Dhole, Du, Durmus, Du{\v{s}}ek, Emezue, Gangal, Garbacea, Hashimoto, Hou, Jernite, Jhamtani, Ji, Jolly, Kale, Kumar, Ladhak, Madaan, Maddela, Mahajan, Mahamood, Majumder, Martins, McMillan-Major, Mille, van Miltenburg, Nadeem, Narayan, Nikolaev, Niyongabo~Rubungo, Osei, Parikh, Perez-Beltrachini, Rao, Raunak, Rodriguez, Santhanam, Sedoc, Sellam, Shaikh, Shimorina, Sobrevilla~Cabezudo, Strobelt, Subramani, Xu, Yang, Yerukola, and Zhou}]{gehrmann-etal-2021-gem}
Sebastian Gehrmann, Tosin Adewumi, Karmanya Aggarwal, Pawan~Sasanka Ammanamanchi, Anuoluwapo Aremu, Antoine Bosselut, Khyathi~Raghavi Chandu, Miruna-Adriana Clinciu, Dipanjan Das, Kaustubh Dhole, Wanyu Du, Esin Durmus, Ond{\v{r}}ej Du{\v{s}}ek, Chris~Chinenye Emezue, Varun Gangal, Cristina Garbacea, Tatsunori Hashimoto, Yufang Hou, Yacine Jernite, Harsh Jhamtani, Yangfeng Ji, Shailza Jolly, Mihir Kale, Dhruv Kumar, Faisal Ladhak, Aman Madaan, Mounica Maddela, Khyati Mahajan, Saad Mahamood, Bodhisattwa~Prasad Majumder, Pedro~Henrique Martins, Angelina McMillan-Major, Simon Mille, Emiel van Miltenburg, Moin Nadeem, Shashi Narayan, Vitaly Nikolaev, Andre Niyongabo~Rubungo, Salomey Osei, Ankur Parikh, Laura Perez-Beltrachini, Niranjan~Ramesh Rao, Vikas Raunak, Juan~Diego Rodriguez, Sashank Santhanam, Jo{\~a}o Sedoc, Thibault Sellam, Samira Shaikh, Anastasia Shimorina, Marco~Antonio Sobrevilla~Cabezudo, Hendrik Strobelt, Nishant Subramani, Wei Xu, Diyi Yang, Akhila Yerukola, and Jiawei Zhou. 2021.
\newblock \href {https://doi.org/10.18653/v1/2021.gem-1.10} {The {GEM} benchmark: Natural language generation, its evaluation and metrics}.
\newblock In \emph{Proceedings of the 1st Workshop on Natural Language Generation, Evaluation, and Metrics (GEM 2021)}, pages 96--120, Online. Association for Computational Linguistics.

\bibitem[{Hardalov et~al.(2020)Hardalov, Mihaylov, Zlatkova, Dinkov, Koychev, and Nakov}]{hardalov-etal-2020-exams}
Momchil Hardalov, Todor Mihaylov, Dimitrina Zlatkova, Yoan Dinkov, Ivan Koychev, and Preslav Nakov. 2020.
\newblock \href {https://doi.org/10.18653/v1/2020.emnlp-main.438} {{EXAMS}: A multi-subject high school examinations dataset for cross-lingual and multilingual question answering}.
\newblock In \emph{Proceedings of the 2020 Conference on Empirical Methods in Natural Language Processing (EMNLP)}, pages 5427--5444, Online. Association for Computational Linguistics.

\bibitem[{Hendrycks et~al.(2021)Hendrycks, Burns, Basart, Zou, Mazeika, Song, and Steinhardt}]{hendrycksmeasuring}
Dan Hendrycks, Collin Burns, Steven Basart, Andy Zou, Mantas Mazeika, Dawn Song, and Jacob Steinhardt. 2021.
\newblock Measuring massive multitask language understanding.
\newblock In \emph{International Conference on Learning Representations}.

\bibitem[{Hosseini et~al.(2021)Hosseini, Reddy, Bahdanau, Hjelm, Sordoni, and Courville}]{DBLP:conf/naacl/HosseiniRBHSC21}
Arian Hosseini, Siva Reddy, Dzmitry Bahdanau, R.~Devon Hjelm, Alessandro Sordoni, and Aaron~C. Courville. 2021.
\newblock \href {https://doi.org/10.18653/v1/2021.naacl-main.102} {Understanding by understanding not: Modeling negation in language models}.
\newblock In \emph{Proceedings of the 2021 Conference of the North American Chapter of the Association for Computational Linguistics: Human Language Technologies, {NAACL-HLT} 2021, Online, June 6-11, 2021}, pages 1301--1312. Association for Computational Linguistics.

\bibitem[{Hu et~al.(2020)Hu, Ruder, Siddhant, Neubig, Firat, and Johnson}]{hu2020xtreme}
Junjie Hu, Sebastian Ruder, Aditya Siddhant, Graham Neubig, Orhan Firat, and Melvin Johnson. 2020.
\newblock \href {http://arxiv.org/abs/arXiv:2003.11080v1} {{XTREME}: A massively multilingual multi-task benchmark for evaluating cross-lingual generalization}.
\newblock In \emph{Proceedings of ICML 2020}.

\bibitem[{Huang et~al.(2023)Huang, Yu, Zhu, Sun, Cheng, Song, Chen, Alharthi, An, Liu et~al.}]{huang2023acegpt}
Huang Huang, Fei Yu, Jianqing Zhu, Xuening Sun, Hao Cheng, Dingjie Song, Zhihong Chen, Abdulmohsen Alharthi, Bang An, Ziche Liu, et~al. 2023.
\newblock Ace{GPT}, localizing large language models in {A}rabic.
\newblock \emph{arXiv preprint arXiv:2309.12053}.

\bibitem[{Huang et~al.(2019)Huang, Le~Bras, Bhagavatula, and Choi}]{huang-etal-2019-cosmos}
Lifu Huang, Ronan Le~Bras, Chandra Bhagavatula, and Yejin Choi. 2019.
\newblock \href {https://doi.org/10.18653/v1/D19-1243} {Cosmos {QA}: Machine reading comprehension with contextual commonsense reasoning}.
\newblock In \emph{Proceedings of the 2019 Conference on Empirical Methods in Natural Language Processing and the 9th International Joint Conference on Natural Language Processing (EMNLP-IJCNLP)}, pages 2391--2401, Hong Kong, China. Association for Computational Linguistics.

\bibitem[{Inoue et~al.(2021)Inoue, Alhafni, Baimukan, Bouamor, and Habash}]{inoue-etal-2021-interplay}
Go~Inoue, Bashar Alhafni, Nurpeiis Baimukan, Houda Bouamor, and Nizar Habash. 2021.
\newblock \href {https://aclanthology.org/2021.wanlp-1.10} {The interplay of variant, size, and task type in {A}rabic pre-trained language models}.
\newblock In \emph{Proceedings of the Sixth Arabic Natural Language Processing Workshop}, pages 92--104, Kyiv, Ukraine (Virtual). Association for Computational Linguistics.

\bibitem[{Kamal~Eddine et~al.(2022)Kamal~Eddine, Tomeh, Habash, Le~Roux, and Vazirgiannis}]{kamal-eddine-etal-2022-arabart}
Moussa Kamal~Eddine, Nadi Tomeh, Nizar Habash, Joseph Le~Roux, and Michalis Vazirgiannis. 2022.
\newblock \href {https://aclanthology.org/2022.wanlp-1.4} {{A}ra{BART}: a pretrained {A}rabic sequence-to-sequence model for abstractive summarization}.
\newblock In \emph{Proceedings of the The Seventh Arabic Natural Language Processing Workshop (WANLP)}, pages 31--42, Abu Dhabi, United Arab Emirates (Hybrid). Association for Computational Linguistics.

\bibitem[{Kassner and Sch{\"{u}}tze(2020)}]{DBLP:conf/acl/KassnerS20}
Nora Kassner and Hinrich Sch{\"{u}}tze. 2020.
\newblock \href {https://doi.org/10.18653/v1/2020.acl-main.698} {Negated and misprimed probes for pretrained language models: Birds can talk, but cannot fly}.
\newblock In \emph{Proceedings of the 58th Annual Meeting of the Association for Computational Linguistics, {ACL} 2020, Online, July 5-10, 2020}, pages 7811--7818. Association for Computational Linguistics.

\bibitem[{Koto et~al.(2023)Koto, Aisyah, Li, and Baldwin}]{koto-etal-2023-large}
Fajri Koto, Nurul Aisyah, Haonan Li, and Timothy Baldwin. 2023.
\newblock \href {https://doi.org/10.18653/v1/2023.emnlp-main.760} {Large language models only pass primary school exams in {I}ndonesia: A comprehensive test on {I}ndo{MMLU}}.
\newblock In \emph{Proceedings of the 2023 Conference on Empirical Methods in Natural Language Processing}, pages 12359--12374, Singapore. Association for Computational Linguistics.

\bibitem[{Koto et~al.(2022)Koto, Baldwin, and Lau}]{koto-etal-2022-cloze}
Fajri Koto, Timothy Baldwin, and Jey~Han Lau. 2022.
\newblock \href {https://doi.org/10.18653/v1/2022.csrr-1.2} {Cloze evaluation for deeper understanding of commonsense stories in {I}ndonesian}.
\newblock In \emph{Proceedings of the First Workshop on Commonsense Representation and Reasoning (CSRR 2022)}, pages 8--16, Dublin, Ireland. Association for Computational Linguistics.

\bibitem[{Koto et~al.(2024)Koto, Mahendra, Aisyah, and Baldwin}]{koto2024indoculture}
Fajri Koto, Rahmad Mahendra, Nurul Aisyah, and Timothy Baldwin. 2024.
\newblock Indo{C}ulture: Exploring geographically-influenced cultural commonsense reasoning across eleven {I}ndonesian provinces.
\newblock \emph{arXiv preprint arXiv:2404.01854}.

\bibitem[{Ladhak et~al.(2020)Ladhak, Durmus, Cardie, and McKeown}]{ladhak-etal-2020-wikilingua}
Faisal Ladhak, Esin Durmus, Claire Cardie, and Kathleen McKeown. 2020.
\newblock \href {https://doi.org/10.18653/v1/2020.findings-emnlp.360} {{W}iki{L}ingua: A new benchmark dataset for cross-lingual abstractive summarization}.
\newblock In \emph{Findings of the Association for Computational Linguistics: EMNLP 2020}, pages 4034--4048, Online. Association for Computational Linguistics.

\bibitem[{Lee et~al.(2023)Lee, Phatale, Mansoor, Mesnard, Ferret, Lu, Bishop, Hall, Carbune, Rastogi, and Prakash}]{rlaif}
Harrison Lee, Samrat Phatale, Hassan Mansoor, Thomas Mesnard, Johan Ferret, Kellie Lu, Colton Bishop, Ethan Hall, Victor Carbune, Abhinav Rastogi, and Sushant Prakash. 2023.
\newblock \href {http://arxiv.org/abs/2309.00267} {{RLAIF}: Scaling reinforcement learning from human feedback with {AI} feedback}.
\newblock arXiv preprint arXiv:2309.00267.

\bibitem[{Lewis et~al.(2020)Lewis, Oguz, Rinott, Riedel, and Schwenk}]{lewis-etal-2020-mlqa}
Patrick Lewis, Barlas Oguz, Ruty Rinott, Sebastian Riedel, and Holger Schwenk. 2020.
\newblock \href {https://doi.org/10.18653/v1/2020.acl-main.653} {{MLQA}: Evaluating cross-lingual extractive question answering}.
\newblock In \emph{Proceedings of the 58th Annual Meeting of the Association for Computational Linguistics}, pages 7315--7330, Online. Association for Computational Linguistics.

\bibitem[{Li et~al.(2023)Li, Zhang, Koto, Yang, Zhao, Gong, Duan, and Baldwin}]{li2023cmmlu}
Haonan Li, Yixuan Zhang, Fajri Koto, Yifei Yang, Hai Zhao, Yeyun Gong, Nan Duan, and Timothy Baldwin. 2023.
\newblock {CMMLU}: Measuring massive multitask language understanding in {C}hinese.
\newblock \emph{arXiv preprint arXiv:2306.09212}.

\bibitem[{Liang et~al.(2020)Liang, Duan, Gong, Wu, Guo, Qi, Gong, Shou, Jiang, Cao, Fan, Zhang, Agrawal, Cui, Wei, Bharti, Qiao, Chen, Wu, Liu, Yang, Campos, Majumder, and Zhou}]{liang-etal-2020-xglue}
Yaobo Liang, Nan Duan, Yeyun Gong, Ning Wu, Fenfei Guo, Weizhen Qi, Ming Gong, Linjun Shou, Daxin Jiang, Guihong Cao, Xiaodong Fan, Ruofei Zhang, Rahul Agrawal, Edward Cui, Sining Wei, Taroon Bharti, Ying Qiao, Jiun-Hung Chen, Winnie Wu, Shuguang Liu, Fan Yang, Daniel Campos, Rangan Majumder, and Ming Zhou. 2020.
\newblock \href {https://doi.org/10.18653/v1/2020.emnlp-main.484} {{XGLUE}: A new benchmark dataset for cross-lingual pre-training, understanding and generation}.
\newblock In \emph{Proceedings of the 2020 Conference on Empirical Methods in Natural Language Processing (EMNLP)}, pages 6008--6018, Online. Association for Computational Linguistics.

\bibitem[{Lin et~al.(2022)Lin, Mihaylov, Artetxe, Wang, Chen, Simig, Ott, Goyal, Bhosale, Du, Pasunuru, Shleifer, Koura, Chaudhary, O{'}Horo, Wang, Zettlemoyer, Kozareva, Diab, Stoyanov, and Li}]{lin-etal-2022-shot}
Xi~Victoria Lin, Todor Mihaylov, Mikel Artetxe, Tianlu Wang, Shuohui Chen, Daniel Simig, Myle Ott, Naman Goyal, Shruti Bhosale, Jingfei Du, Ramakanth Pasunuru, Sam Shleifer, Punit~Singh Koura, Vishrav Chaudhary, Brian O{'}Horo, Jeff Wang, Luke Zettlemoyer, Zornitsa Kozareva, Mona Diab, Veselin Stoyanov, and Xian Li. 2022.
\newblock \href {https://aclanthology.org/2022.emnlp-main.616} {Few-shot learning with multilingual generative language models}.
\newblock In \emph{Proceedings of the 2022 Conference on Empirical Methods in Natural Language Processing}, pages 9019--9052, Abu Dhabi, United Arab Emirates. Association for Computational Linguistics.

\bibitem[{Liu et~al.(2023)Liu, Qiao, Neiswanger, Wang, Tan, Tao, Li, Wang, Sun, Pangarkar, Fan, Gu, Miller, Zhuang, He, Li, Koto, Tang, Ranjan, Shen, Ren, Iriondo, Mu, Hu, Schulze, Nakov, Baldwin, and Xing}]{liu2023llm360}
Zhengzhong Liu, Aurick Qiao, Willie Neiswanger, Hongyi Wang, Bowen Tan, Tianhua Tao, Junbo Li, Yuqi Wang, Suqi Sun, Omkar Pangarkar, Richard Fan, Yi~Gu, Victor Miller, Yonghao Zhuang, Guowei He, Haonan Li, Fajri Koto, Liping Tang, Nikhil Ranjan, Zhiqiang Shen, Xuguang Ren, Roberto Iriondo, Cun Mu, Zhiting Hu, Mark Schulze, Preslav Nakov, Timothy Baldwin, and Eric~P. Xing. 2023.
\newblock \href {https://api.semanticscholar.org/CorpusID:266162750} {{LLM}360: Towards fully transparent open-source {LLM}s}.
\newblock \emph{ArXiv}, abs/2312.06550.

\bibitem[{Mozannar et~al.(2019)Mozannar, Maamary, El~Hajal, and Hajj}]{mozannar-etal-2019-neural}
Hussein Mozannar, Elie Maamary, Karl El~Hajal, and Hazem Hajj. 2019.
\newblock \href {https://doi.org/10.18653/v1/W19-4612} {Neural {A}rabic question answering}.
\newblock In \emph{Proceedings of the Fourth Arabic Natural Language Processing Workshop}, pages 108--118, Florence, Italy. Association for Computational Linguistics.

\bibitem[{Muennighoff et~al.(2022)Muennighoff, Wang, Sutawika, Roberts, Biderman, Scao, Bari, Shen, Yong, Schoelkopf et~al.}]{muennighoff2022crosslingual}
Niklas Muennighoff, Thomas Wang, Lintang Sutawika, Adam Roberts, Stella Biderman, Teven~Le Scao, M~Saiful Bari, Sheng Shen, Zheng-Xin Yong, Hailey Schoelkopf, et~al. 2022.
\newblock Crosslingual generalization through multitask finetuning.
\newblock \emph{arXiv preprint arXiv:2211.01786}.

\bibitem[{Nagoudi et~al.(2022)Nagoudi, Elmadany, and Abdul-Mageed}]{nagoudi-etal-2022-arat5}
El~Moatez~Billah Nagoudi, AbdelRahim Elmadany, and Muhammad Abdul-Mageed. 2022.
\newblock \href {https://doi.org/10.18653/v1/2022.acl-long.47} {{A}ra{T}5: Text-to-text transformers for {A}rabic language generation}.
\newblock In \emph{Proceedings of the 60th Annual Meeting of the Association for Computational Linguistics (Volume 1: Long Papers)}, pages 628--647, Dublin, Ireland. Association for Computational Linguistics.

\bibitem[{Nagoudi et~al.(2023)Nagoudi, Elmadany, El-Shangiti, and Abdul-Mageed}]{nagoudi-etal-2023-dolphin}
El~Moatez~Billah Nagoudi, AbdelRahim Elmadany, Ahmed El-Shangiti, and Muhammad Abdul-Mageed. 2023.
\newblock \href {https://doi.org/10.18653/v1/2023.findings-emnlp.98} {Dolphin: A challenging and diverse benchmark for {A}rabic {NLG}}.
\newblock In \emph{Findings of the Association for Computational Linguistics: EMNLP 2023}, pages 1404--1422, Singapore. Association for Computational Linguistics.

\bibitem[{OpenAI(2023)}]{OpenAI2023GPT4TR}
OpenAI. 2023.
\newblock {GPT-4} technical report.
\newblock \emph{ArXiv}, abs/2303.08774.

\bibitem[{Ouyang et~al.(2022)Ouyang, Wu, Jiang, Almeida, Wainwright, Mishkin, Zhang, Agarwal, Slama, Ray et~al.}]{ouyang2022training}
Long Ouyang, Jeffrey Wu, Xu~Jiang, Diogo Almeida, Carroll Wainwright, Pamela Mishkin, Chong Zhang, Sandhini Agarwal, Katarina Slama, Alex Ray, et~al. 2022.
\newblock Training language models to follow instructions with human feedback.
\newblock \emph{Advances in Neural Information Processing Systems}, 35:27730--27744.

\bibitem[{Pan et~al.(2017)Pan, Zhang, May, Nothman, Knight, and Ji}]{pan-etal-2017-cross}
Xiaoman Pan, Boliang Zhang, Jonathan May, Joel Nothman, Kevin Knight, and Heng Ji. 2017.
\newblock \href {https://doi.org/10.18653/v1/P17-1178} {Cross-lingual name tagging and linking for 282 languages}.
\newblock In \emph{Proceedings of the 55th Annual Meeting of the Association for Computational Linguistics (Volume 1: Long Papers)}, pages 1946--1958, Vancouver, Canada. Association for Computational Linguistics.

\bibitem[{Penedo et~al.(2023)Penedo, Malartic, Hesslow, Cojocaru, Cappelli, Alobeidli, Pannier, Almazrouei, and Launay}]{penedo2023refinedweb}
Guilherme Penedo, Quentin Malartic, Daniel Hesslow, Ruxandra Cojocaru, Alessandro Cappelli, Hamza Alobeidli, Baptiste Pannier, Ebtesam Almazrouei, and Julien Launay. 2023.
\newblock The {RefinedWeb} dataset for {Falcon LLM}: Outperforming curated corpora with web data, and web data only.
\newblock \emph{arXiv preprint arXiv:2306.01116}.

\bibitem[{Ramesh et~al.(2023)Ramesh, Sitaram, and Choudhury}]{ramesh-etal-2023-fairness}
Krithika Ramesh, Sunayana Sitaram, and Monojit Choudhury. 2023.
\newblock \href {https://aclanthology.org/2023.findings-eacl.157} {Fairness in language models beyond {E}nglish: Gaps and challenges}.
\newblock In \emph{Findings of the Association for Computational Linguistics: EACL 2023}, pages 2106--2119, Dubrovnik, Croatia. Association for Computational Linguistics.

\bibitem[{Ruder et~al.(2021)Ruder, Constant, Botha, Siddhant, Firat, Fu, Liu, Hu, Garrette, Neubig, and Johnson}]{ruder-etal-2021-xtreme}
Sebastian Ruder, Noah Constant, Jan Botha, Aditya Siddhant, Orhan Firat, Jinlan Fu, Pengfei Liu, Junjie Hu, Dan Garrette, Graham Neubig, and Melvin Johnson. 2021.
\newblock \href {https://doi.org/10.18653/v1/2021.emnlp-main.802} {{XTREME}-{R}: Towards more challenging and nuanced multilingual evaluation}.
\newblock In \emph{Proceedings of the 2021 Conference on Empirical Methods in Natural Language Processing}, pages 10215--10245, Online and Punta Cana, Dominican Republic. Association for Computational Linguistics.

\bibitem[{Sengupta et~al.(2023)Sengupta, Sahu, Jia, Katipomu, Li, Koto, Afzal, Kamboj, Pandit, Pal et~al.}]{sengupta2023jais}
Neha Sengupta, Sunil~Kumar Sahu, Bokang Jia, Satheesh Katipomu, Haonan Li, Fajri Koto, Osama~Mohammed Afzal, Samta Kamboj, Onkar Pandit, Rahul Pal, et~al. 2023.
\newblock Jais and {Jais-chat}: {Arabic}-centric foundation and instruction-tuned open generative large language models.
\newblock \emph{arXiv preprint arXiv:2308.16149}.

\bibitem[{Shoufan and Alameri(2015)}]{shoufan-alameri-2015-natural}
Abdulhadi Shoufan and Sumaya Alameri. 2015.
\newblock \href {https://doi.org/10.18653/v1/W15-3205} {Natural language processing for dialectical {A}rabic: A survey}.
\newblock In \emph{Proceedings of the Second Workshop on {A}rabic Natural Language Processing}, pages 36--48, Beijing, China. Association for Computational Linguistics.

\bibitem[{Talat et~al.(2022)Talat, N{\'e}v{\'e}ol, Biderman, Clinciu, Dey, Longpre, Luccioni, Masoud, Mitchell, Radev, Sharma, Subramonian, Tae, Tan, Tunuguntla, and Van Der~Wal}]{talat-etal-2022-reap}
Zeerak Talat, Aur{\'e}lie N{\'e}v{\'e}ol, Stella Biderman, Miruna Clinciu, Manan Dey, Shayne Longpre, Sasha Luccioni, Maraim Masoud, Margaret Mitchell, Dragomir Radev, Shanya Sharma, Arjun Subramonian, Jaesung Tae, Samson Tan, Deepak Tunuguntla, and Oskar Van Der~Wal. 2022.
\newblock \href {https://doi.org/10.18653/v1/2022.bigscience-1.3} {You reap what you sow: On the challenges of bias evaluation under multilingual settings}.
\newblock In \emph{Proceedings of BigScience Episode {\#}5 -- Workshop on Challenges {\&} Perspectives in Creating Large Language Models}, pages 26--41, virtual+Dublin. Association for Computational Linguistics.

\bibitem[{Touvron et~al.(2023)Touvron, Martin, Stone, Albert, Almahairi, Babaei, Bashlykov, Batra, Bhargava, Bhosale et~al.}]{touvron2023llama2}
Hugo Touvron, Louis Martin, Kevin Stone, Peter Albert, Amjad Almahairi, Yasmine Babaei, Nikolay Bashlykov, Soumya Batra, Prajjwal Bhargava, Shruti Bhosale, et~al. 2023.
\newblock Llama2: Open foundation and fine-tuned chat models.
\newblock \emph{arXiv preprint arXiv:2307.09288}.

\bibitem[{Truong et~al.(2023)Truong, Baldwin, Verspoor, and Cohn}]{truong-etal-2023-language}
Thinh~Hung Truong, Timothy Baldwin, Karin Verspoor, and Trevor Cohn. 2023.
\newblock \href {https://doi.org/10.18653/v1/2023.starsem-1.10} {Language models are not naysayers: an analysis of language models on negation benchmarks}.
\newblock In \emph{Proceedings of the 12th Joint Conference on Lexical and Computational Semantics (*SEM 2023)}, pages 101--114, Toronto, Canada. Association for Computational Linguistics.

\bibitem[{Wolf et~al.(2020)Wolf, Debut, Sanh, Chaumond, Delangue, Moi, Cistac, Rault, Louf, Funtowicz, Davison, Shleifer, von Platen, Ma, Jernite, Plu, Xu, Le~Scao, Gugger, Drame, Lhoest, and Rush}]{wolf-etal-2020-transformers}
Thomas Wolf, Lysandre Debut, Victor Sanh, Julien Chaumond, Clement Delangue, Anthony Moi, Pierric Cistac, Tim Rault, Remi Louf, Morgan Funtowicz, Joe Davison, Sam Shleifer, Patrick von Platen, Clara Ma, Yacine Jernite, Julien Plu, Canwen Xu, Teven Le~Scao, Sylvain Gugger, Mariama Drame, Quentin Lhoest, and Alexander Rush. 2020.
\newblock \href {https://doi.org/10.18653/v1/2020.emnlp-demos.6} {Transformers: State-of-the-art natural language processing}.
\newblock In \emph{Proceedings of the 2020 Conference on Empirical Methods in Natural Language Processing: System Demonstrations}, pages 38--45, Online. Association for Computational Linguistics.

\bibitem[{Yu et~al.(2024)Yu, Shen, Ran, Zhang, Zhang, Ma, Liang, Li, Wang, and Xie}]{yu2024codereval}
Hao Yu, Bo~Shen, Dezhi Ran, Jiaxin Zhang, Qi~Zhang, Yuchi Ma, Guangtai Liang, Ying Li, Qianxiang Wang, and Tao Xie. 2024.
\newblock Coder{E}val: A benchmark of pragmatic code generation with generative pre-trained models.
\newblock In \emph{Proceedings of the 46th IEEE/ACM International Conference on Software Engineering}, pages 1--12.

\bibitem[{Zellers et~al.(2019)Zellers, Holtzman, Bisk, Farhadi, and Choi}]{zellers-etal-2019-hellaswag}
Rowan Zellers, Ari Holtzman, Yonatan Bisk, Ali Farhadi, and Yejin Choi. 2019.
\newblock \href {https://doi.org/10.18653/v1/P19-1472} {{H}ella{S}wag: Can a machine really finish your sentence?}
\newblock In \emph{Proceedings of the 57th Annual Meeting of the Association for Computational Linguistics}, pages 4791--4800, Florence, Italy. Association for Computational Linguistics.

\end{thebibliography}
\bibliographystyle{acl_natbib}

\clearpage

\appendix

\section{Data Statistics}
Table~\ref{tab:stat_per_subject} presents the distribution of ArabicMMLU data categorized by subject across different education levels. 
\begin{table}[ht!]
    \centering
    \resizebox{0.8\linewidth}{!}{
        \begin{tabular}{L{5cm}r}
        \toprule
        \textbf{Subject} &  \textbf{\#question} \\        
        \midrule
        \multicolumn{2}{l}{\textbf{Primary school}} \\
        Arabic Language  & 255 \\
        Computer Science &  193 \\
        General Knowledge &  165 \\
        Geography &  60 \\
        History &   105 \\
        Islamic Studies & 1002 \\
        Math &   412 \\
        Natural Science &   339 \\
        Social Science &   708 \\
        \midrule
        \multicolumn{2}{l}{\textbf{Middle school}} \\   
        Arabic Language & 30 \\
        Civics & 239 \\
        Computer Science & 30 \\
        Economics & 90 \\
        General Knowledge & 175 \\
        Geography & 275 \\
        History & 206 \\
        Islamic Studies & 241 \\
        Natural Science & 245 \\
        Social Science & 244 \\
        \midrule
        \multicolumn{2}{l}{\textbf{High school}} \\   
        Arabic Language & 393 \\
        Biology & 1412 \\
        Civics & 90 \\
        Computer Science & 264 \\
        Economics & 363 \\
        Geography & 1041 \\
        History & 763 \\
        Islamic Studies & 337 \\
        Philosophy & 42 \\
        Physics & 258 \\
        \midrule
        \multicolumn{2}{l}{\textbf{University and Professional}} \\
        Accounting & 77 \\
        Computer Science & 67 \\
        Economics & 140 \\
        Management & 78 \\ 
        Political Science & 213 \\
        Law & 317 \\
        \midrule
        \multicolumn{2}{l}{\textbf{Other / NA}} \\   
        Arabic Language (General) & 615 \\
        Arabic Language (Grammar) & 368 \\
        Driving Test & 1214 \\ 
        General Knowledge & 867 \\
        Islamic Studies & 642 \\
        \midrule
        \textbf{Total} & \textbf{14575} \\
        \bottomrule
        \end{tabular}
    }
    \caption{The distribution of \datasetname{} for each subject in different education levels.}
    \label{tab:stat_per_subject}
\end{table}

\newpage

\section{Examples}
\label{app:examples}
\begin{figure}[ht!]
    \centering
    \includegraphics[width=\linewidth]{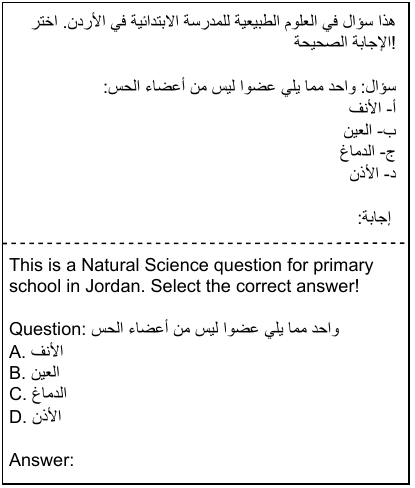} 
    \caption{Example of prompt input in Arabic and English.}
    \label{fig:prompt_example}
\end{figure}

Figure~\ref{fig:prompt_example} illustrates a complete example of prompts used in this study. This example features a Natural Science question with prompts provided in both Arabic and English.

\onecolumn

\section{Detailed Experiment Results}
Table~\ref{tab:en_en_detailed} presents the detailed zero-shot results across subjects and education levels, while Table~\ref{tab:result_ar_ar}, Table~\ref{tab:result_ar_en}, Table~\ref{tab:result_en_ar} display the results with different prompts and alphabetic outputs (complementing the main result at Table~\ref{tab:en_en_detailed}).

\begin{table*}[ht!]
    \centering
    \resizebox{\linewidth}{!}{
        \begin{tabular}{L{4.5cm}R{2.5cm}R{2.5cm}R{2.5cm}R{2.5cm}R{2.5cm}}
        \toprule
        \textbf{Subject} & \textbf{BLOOMZ} & \textbf{AceGPT-chat} & \textbf{Jais-chat} & \textbf{GPT-3.5} & \textbf{GPT-4} \\
        \midrule
        \multicolumn{6}{l}{\textbf{Primary School}} \\
        Arabic Language & 64.3 & 60.7 & 63.1 & 65.1 & 80.6 \\
        Computer Science & 62.6 & 65.3 & 68.9 & 66.8 & 80.5 \\
        General Knowledge & 62.3 & 66.0 & 74.7 & 75.9 & 77.2 \\
        Geography & 50.9 & 57.9 & 61.4 & 66.7 & 82.5 \\
        History & 48.0 & 52.0 & 75.5 & 56.9 & 71.6 \\
        Islamic Studies & 67.0 & 71.6 & 81.8 & 74.0 & 89.8 \\
        Math & 41.3 & 48.9 & 57.9 & 58.2 & 76.0 \\
        Natural Science & 67.3 & 68.5 & 82.1 & 80.4 & 88.7 \\
        Social Science & 62.7 & 69.2 & 75.7 & 74.3 & 84.7 \\
        \midrule
        \multicolumn{6}{l}{\textbf{Middle School}} \\
        Arabic Language & 51.9 & 51.9 & 77.8 & 55.6 & 85.2 \\
        Civics & 40.3 & 40.3 & 60.2 & 45.3 & 62.7 \\
        Computer Science & 88.9 & 74.1 & 85.2 & 81.5 & 96.3 \\
        Economics & 72.4 & 66.7 & 77.0 & 77.0 & 81.6 \\
        General Knowledge & 59.0 & 65.3 & 70.5 & 67.6 & 78.6 \\
        Geography & 50.7 & 57.7 & 67.3 & 62.5 & 75.4 \\
        History & 52.7 & 61.1 & 68.5 & 62.6 & 71.9 \\
        Islamic Studies & 56.7 & 55.5 & 73.1 & 62.6 & 73.9 \\
        Natural Science & 51.7 & 61.6 & 69.8 & 70.2 & 87.2 \\
        Social Science & 42.7 & 49.4 & 54.4 & 49.8 & 57.7 \\
        \midrule
        \multicolumn{6}{l}{\textbf{High School}} \\
        Arabic Language & 33.8 & 35.6 & 44.6 & 36.7 & 44.6 \\
        Biology & 35.0 & 37.6 & 46.7 & 42.4 & 59.6 \\
        Civics & 39.1 & 36.8 & 47.7 & 39.1 & 44.8 \\
        Computer Science & 42.1 & 52.1 & 55.6 & 57.9 & 74.7 \\
        Economics & 45.8 & 48.9 & 58.1 & 56.7 & 71.1 \\
        Geography & 40.2 & 46.3 & 53.1 & 49.0 & 66.1 \\
        History & 38.9 & 40.5 & 50.6 & 42.7 & 54.1 \\
        Islamic Studies & 52.8 & 51.3 & 66.9 & 62.4 & 76.7 \\
        Philosophy & 59.0 & 53.8 & 66.7 & 59.0 & 74.4 \\
        Physics & 32.5 & 34.1 & 43.9 & 42.0 & 61.6 \\
        \midrule
        \multicolumn{6}{l}{\textbf{University and Professional}} \\
        Accounting & 50.0 & 55.4 & 55.4 & 59.5 & 73.0 \\
        Computer Science & 48.4 & 53.1 & 67.2 & 57.8 & 78.1 \\
        Economics & 48.9 & 43.8 & 52.6 & 52.6 & 62.8 \\
        Management & 48.7 & 65.8 & 78.9 & 64.5 & 80.3 \\
        Political Science & 44.3 & 52.9 & 54.8 & 51.4 & 66.7 \\
        Law & 25.9 & 52.7 & 33.1 & 55.8 & 66.9 \\
        \midrule
        \multicolumn{6}{l}{\textbf{Other / NA}} \\
        Arabic Language (General) & 58.5 & 57.8 & 72.7 & 66.7 & 84.5 \\
        Arabic Language (Grammar) & 42.5 & 46.8 & 60.5 & 59.7 & 77.3 \\
        Driving Test & 52.3 & 61.8 & 65.9 & 68.3 & 79.5 \\
        General Knowledge & 42.5 & 50.4 & 68.9 & 54.5 & 72.5 \\
        Islamic Studies & 38.7 & 41.9 & 67.4 & 44.0 & 71.8 \\
        \bottomrule
        \end{tabular}
    }
      \caption{Zero-shot LLM performance (\% accuracy) with \textbf{English prompt and English alphabetic output}, for each subject and education level. The models are BLOOMZ (7B), AceGPT-chat (13B), Jais-chat (30B), GPT-3.5 (175B), and GPT-4.}
    \label{tab:en_en_detailed}
\end{table*}

\begin{table*}[t]
    \centering
    \resizebox{\linewidth}{!}{
        \begin{tabular}{lC{2cm}C{2cm}C{2cm}C{2cm}C{2cm}C{2cm}}
        \toprule
       \multirow{2}{*}{\textbf{Model (\#parameters)}}& \multirow{2}{*}{\textbf{STEM}} & \textbf{Social} & \multirow{2}{*}{\textbf{Humanities}} & \textbf{Arabic} &  \multirow{2}{*}{\textbf{Other}}  & \multirow{2}{*}{\textbf{Average}} \\
       & & \textbf{Science} & & \textbf{Language} & & \\
       \midrule
        Random & 29.5 & 28.9 & 28.6 & 25.8 & 32.3 & 29.0 \\
        \hdashline
        XGLM (1.7B) & 28.8 & 28.0 & 25.4 & 25.1 & 28.7 & 27.3 \\
        XGLM (2.9B) & 28.8 & 26.9 & 26.8 & 26.5 & 32.3 & 28.2 \\
        XGLM (4.5B) & 30.5 & 27.0 & 27.1 & 26.6 & 32.0 & 28.6 \\
        XGLM (7.5B) & 30.3 & 25.7 & 25.8 & 25.0 & 32.6 & 27.9 \\
        \hdashline
        BLOOMZ (560M) & 29.3 & 26.3 & 27.1 & 23.7 & 27.2 & 27.0 \\
        BLOOMZ (1.1B) & 31.3 & 28.1 & 31.0 & 28.3 & 29.0 & 29.7 \\
        BLOOMZ (1.7B) & 32.5 & 34.9 & 35.2 & 30.4 & 35.1 & 34.0 \\
        BLOOMZ (3B) & 38.3 & 42.6 & 40.0 & 36.2 & 39.5 & 39.7 \\
        BLOOMZ (7B) & 37.7 & 40.5 & 34.8 & 38.2 & 39.6 & 38.0 \\
        \hdashline
        mT0$_\text{small}$ (300M) & 29.1 & 28.7 & 26.0 & 22.5 & 27.3 & 27.2 \\
        mT0$_\text{base}$ (580M) & 30.2 & 30.5 & 33.1 & 24.8 & 34.3 & 31.1 \\
        mT0$_\text{large}$ (1.2B) & 29.4 & 28.8 & 23.9 & 22.7 & 27.2 & 26.7 \\
        mT0$_\text{xl}$ (3.7B) & 39.0 & 40.2 & 39.5 & 41.3 & 43.7 & 40.5 \\
        mT0$_\text{xxl}$ (13B ) & 40.3 & 43.5 & 41.3 & 38.6 & 43.3 & 41.7 \\
        \hdashline
        LLaMA2 (7B) & 31.7 & 31.3 & 33.2 & 27.2 & 32.2 & 31.6 \\
        LLaMA2-chat (7B) & 31.5 & 31.4 & 30.9 & 26.6 & 30.9 & 30.7 \\
        LLaMA2 (13B) & 31.8 & 31.7 & 32.5 & 29.3 & 38.4 & 32.8 \\
        LLaMA2-chat (13B) & 30.8 & 30.2 & 25.4 & 24.7 & 29.7 & 28.4 \\
        \hdashline
        Falcon (7B) & 29.3 & 27.6 & 26.8 & 23.8 & 28.1 & 27.4 \\
        Falcon-instruct (7B) & 28.9 & 28.7 & 26.5 & 22.3 & 27.5 & 27.3 \\
        Falcon (40B) & 30.1 & 30.3 & 31.1 & 24.8 & 31.5 & 30.0 \\
        Falcon-instruct (40B) & 29.3 & 29.0 & 27.6 & 22.9 & 28.2 & 27.9 \\
        \hdashline
        AraT5 (220M) & 28.2 & 25.7 & 23.5 & 24.2 & 26.7 & 25.7 \\
        AraT5v2 (220M) & 31.2 & 31.1 & 33.0 & 27.7 & 34.5 & 31.8 \\
        AraGPT2 (1.7B) & 29.9 & 30.5 & 31.6 & 28.1 & 35.1 & 31.2\\
        \hdashline
        AceGPT (7B) & 31.8 & 28.2 & 29.9 & 28.0 & 31.9 & 30.0 \\
        AceGPT-chat (7B) & 42.9 & 47.7 & 50.5 & 42.6 & 52.5 & 47.6 \\ 
        AceGPT (13B) & 38.4 & 42.0 & 42.1 & 36.8 & 41.8 & 40.6 \\
        AceGPT-chat (13B) & 44.3 & 50.9 & 49.0 & 50.8 & 53.8 & 49.4 \\
        \hdashline
        Jais (13B) & 31.6 & 34.4 & 35.9 & 29.7 & 38.5 & 34.3 \\
        Jais-chat (13B) & 51.6 & 55.1 & 57.3 & 41.1 & 59.3 & 54.0 \\
        Jais (30B) & 33.2 & 35.1 & 34.4 & 27.7 & 39.4 & 34.4 \\
        Jais-chat (30B) & \textbf{53.3} & \textbf{57.9} & \textbf{62.9} & \textbf{60.0} & \textbf{66.8} & \textbf{59.9} \\
        \bottomrule
        \end{tabular}
    }
    \caption{Zero-shot LLM performance (\% accuracy) with \textbf{Arabic prompt and Arabic alphabetic output}, combined across subject groups. ``Average'' means the average across all questions in \datasetname{}.}
    \label{tab:result_ar_ar}
\end{table*}

\begin{table*}[t]
    \centering
    \resizebox{\linewidth}{!}{
        \begin{tabular}{lC{2cm}C{2cm}C{2cm}C{2cm}C{2cm}C{2cm}}
        \toprule
       \multirow{2}{*}{\textbf{Model (\#parameters)}}& \multirow{2}{*}{\textbf{STEM}} & \textbf{Social} & \multirow{2}{*}{\textbf{Humanities}} & \textbf{Arabic} &  \multirow{2}{*}{\textbf{Other}}  & \multirow{2}{*}{\textbf{Average}} \\
       & & \textbf{Science} & & \textbf{Language} & & \\
       \midrule
        Random & 29.5 & 28.9 & 28.6 & 25.8 & 32.3 & 29.0 \\
        \hdashline
        XGLM (1.7B) & 29.9 & 30.7 & 30.8 & 27.7 & 34.8 & 30.9 \\
        XGLM (2.9B) & 29.4 & 30.7 & 31.2 & 27.9 & 34.4 & 30.9 \\
        XGLM (4.5B) & 28.8 & 29.8 & 30.5 & 27.4 & 31.5 & 29.8 \\
        XGLM (7.5B) & 27.7 & 27.5 & 24.8 & 26.5 & 29.3 & 27.0 \\
        \hdashline
        BLOOMZ (560M) & 31.2 & 30.9 & 33.1 & 28.1 & 35.7 & 32.0 \\
        BLOOMZ (1.1B) & 30.3 & 26.7 & 31.1 & 25.5 & 27.5 & 28.6 \\
        BLOOMZ (1.7B) & 36.3 & 38.8 & 38.7 & 38.0 & 39.1 & 38.2 \\
        BLOOMZ (3B) & 40.5 & 45.5 & 44.3 & 43.9 & 47.8 & 44.3 \\
        BLOOMZ (7B) & 43.3 & 47.4 & 47.5 & 50.9 & 50.4 & 47.4 \\
        \hdashline
        mT0$_\text{small}$ (300M) & 30.7 & 30.7 & 31.4 & 28.0 & 34.5 & 31.2 \\
        mT0$_\text{base}$ (580M) & 30.6 & 31.0 & 31.6 & 29.3 & 35.7 & 31.7 \\
        mT0$_\text{large}$ (1.2B) & 30.0 & 30.0 & 29.9 & 29.2 & 34.6 & 30.7 \\
        mT0$_\text{xl}$ (3.7B) & 39.5 & 43.9 & 40.9 & 43.4 & 43.9 & 42.1 \\
        mT0$_\text{xxl}$ (13B ) & 41.2 & 45.2 & 43.3 & 46.7 & 43.8 & 43.8 \\
        \hdashline
        LLaMA2 (7B) & 32.2 & 29.0 & 31.4 & 27.2 & 30.3 & 30.3 \\
        LLaMA2-chat (7B) & 31.7 & 30.7 & 29.5 & 30.3 & 31.4 & 30.7 \\
        LLaMA2 (13B) & 33.2 & 34.1 & 35.3 & 31.7 & 39.5 & 34.9 \\
        LLaMA2-chat (13B) & 33.3 & 30.8 & 30.7 & 31.5 & 36.2 & 32.4 \\
        \hdashline
        Falcon (7B) & 29.8 & 30.6 & 31.4 & 28.2 & 35.1 & 31.1 \\
        Falcon-instruct (7B) & 29.9 & 30.7 & 31.5 & 28.0 & 35.1 & 31.2 \\
        Falcon (40B) & 34.8 & 31.8 & 34.3 & 29.9 & 38.6 & 34.1 \\
        Falcon-instruct (40B) & 33.3 & 29.3 & 33.3 & 30.9 & 39.3 & 33.1 \\
        \hdashline
        AraT5 (220M) & 29.9 & 30.3 & 33.0 & 28.5 & 32.0 & 31.0 \\ 
        AraT5v2 (220M) & 29.1 & 29.8 & 31.1 & 28.3 & 33.6 & 30.5 \\
        AraGPT2 (1.7B) & 33.0 & 31.5 & 35.8 & 29.6 & 37.4 & 33.7 \\
        \hdashline
        AceGPT (7B) & 33.6 & 32.3 & 35.2 & 27.6 & 38.9 & 33.9 \\
        AceGPT-chat (7B) & 42.4 & 47.2 & 49.8 & 41.4 & 51.3 & 46.9 \\
        AceGPT (13B) & 43.2 & 46.6 & 47.5 & 42.4 & 50.0 & 46.2 \\
        AceGPT-chat (13B) & 46.7 & 53.2 & 52.8 & 50.7 & 57.3 & 52.1 \\
        \hdashline
        Jais (13B) & 32.5 & 35.1 & 34.3 & 28.2 & 37.4 & 33.9 \\
        Jais-chat (13B) & 52.4 & 56.6 & 60.0 & 42.5 & 60.4 & 55.6 \\
        Jais (30B) & 39.6 & 45.1 & 49.0 & 32.9 & 49.1 & 44.2 \\
        Jais-chat (30B) & \textbf{55.7} & \textbf{59.7} & \textbf{67.5} & \textbf{61.4} & \textbf{68.3} & \textbf{62.4} \\
        \bottomrule
        \end{tabular}
    }
    \caption{Zero-shot LLM performance (\% accuracy) with \textbf{Arabic prompt and English alphabetic output}, combined across subject groups. ``Average'' means the average across all questions in \datasetname{}.}
    \label{tab:result_ar_en}
\end{table*}

\begin{table*}[t]
    \centering
    \resizebox{\linewidth}{!}{
        \begin{tabular}{lC{2cm}C{2cm}C{2cm}C{2cm}C{2cm}C{2cm}}
        \toprule
       \multirow{2}{*}{\textbf{Model (\#parameters)}}& \multirow{2}{*}{\textbf{STEM}} & \textbf{Social} & \multirow{2}{*}{\textbf{Humanities}} & \textbf{Arabic} &  \multirow{2}{*}{\textbf{Other}}  & \multirow{2}{*}{\textbf{Average}} \\
       & & \textbf{Science} & & \textbf{Language} & & \\
       \midrule
        Random & 29.5 & 28.9 & 28.6 & 25.8 & 32.3 & 29.0 \\
        \hdashline
        XGLM (1.7B) & 30.0 & 29.9 & 26.7 & 27.2 & 29.6 & 28.7 \\
        XGLM (2.9B) & 29.1 & 27.2 & 29.5 & 27.8 & 31.0 & 28.9 \\
        XGLM (4.5B) & 29.8 & 26.8 & 26.9 & 27.6 & 31.8 & 28.4 \\
        XGLM (7.5B) & 30.4 & 26.3 & 26.7 & 27.8 & 32.4 & 28.5 \\
        \hdashline
        BLOOMZ (560M) & 29.5 & 25.9 & 26.3 & 23.9 & 27.1 & 26.8 \\
        BLOOMZ (1.1B) & 31.3 & 29.3 & 30.4 & 28.3 & 29.3 & 29.9 \\
        BLOOMZ (1.7B) & 32.0 & 33.5 & 33.6 & 30.0 & 34.3 & 32.9 \\
        BLOOMZ (3B) & 39.3 & 42.0 & 41.8 & 35.2 & 40.9 & 40.4 \\
        BLOOMZ (7B) & 37.6 & 41.3 & 36.2 & 38.3 & 40.8 & 38.8 \\
        \hdashline
        mT0$_\text{small}$ (300M) & 29.1 & 28.4 & 27.0 & 22.7 & 27.5 & 27.4 \\
        mT0$_\text{base}$ (580M) & 29.5 & 30.3 & 33.3 & 25.3 & 32.6 & 30.7 \\
        mT0$_\text{large}$ (1.2B) & 28.6 & 28.3 & 24.6 & 22.7 & 27.3 & 26.6 \\
        mT0$_\text{xl}$ (3.7B) & 36.8 & 38.9 & 37.7 & 39.8 & 43.2 & 39.0 \\
        mT0$_\text{xxl}$ (13B ) & 39.1 & 41.9 & 40.0 & 36.7 & 42.1 & 40.2 \\
        \hdashline
        LLaMA2 (7B) & 33.0 & 31.2 & 35.5 & 29.5 & 34.4 & 33.0 \\
        LLaMA2-chat (7B) & 34.5 & 33.1 & 31.3 & 27.7 & 34.9 & 32.6 \\
        LLaMA2 (13B) & 33.5 & 30.9 & 31.7 & 30.6 & 35.0 & 32.3 \\
        LLaMA2-chat (13B) & 34.8 & 33.6 & 31.7 & 28.7 & 36.6 & 33.3 \\
        \hdashline
        Falcon (7B) & 29.9 & 30.3 & 34.4 & 27.7 & 32.5 & 31.3 \\
        Falcon-instruct (7B) & 28.5 & 28.4 & 28.9 & 23.0 & 27.8 & 27.8 \\
        Falcon (40B) & 32.4 & 31.6 & 34.7 & 26.9 & 33.3 & 32.3 \\
        Falcon-instruct (40B) & 30.3 & 31.2 & 29.5 & 23.5 & 29.4 & 29.4 \\
        \hdashline
        AraT5 (220M) & 28.1 & 25.7 & 23.4 & 24.8 & 26.7 & 25.7 \\
        AraT5v2 (220M) & 31.3 & 30.0 & 32.9 & 27.1 & 32.9 & 31.2 \\
        AraGPT2 (1.7B) & 29.9 & 30.5 & 31.6 & 28.1 & 35.1 & 31.2 \\
        \hdashline
        AceGPT (7B) & 28.6 & 26.5 & 25.7 & 26.1 & 27.7 & 26.9 \\
        AceGPT-chat (7B) & 43.0 & 46.5 & 49.4 & 42.8 & 52.2 & 47.0 \\
        AceGPT (13B) & 37.6 & 38.9 & 40.1 & 34.3 & 43.6 & 39.2 \\
        AceGPT-chat (13B) & 46.4 & 50.9 & 50.1 & 50.2 & 54.7 & 50.3 \\
        \hdashline
        Jais (13B) & 30.5 & 32.0 & 34.5 & 28.7 & 36.3 & 32.7 \\
        Jais-chat (13B) & 49.2 & 53.4 & 55.8 & 38.5 & 59.5 & 52.4 \\
        Jais (30B) & 39.1 & 43.0 & 47.5 & 32.9 & 49.1 & 43.2 \\
        Jais-chat (30B) & \textbf{54.7} & \textbf{58.8} & \textbf{63.3} & \textbf{59.7} & \textbf{67.4} & \textbf{60.6} \\
        \bottomrule
        \end{tabular}
    }
    \caption{Zero-shot LLM performance (\% accuracy) with \textbf{English prompt and Arabic alphabetic output}, combined across subject groups. ``Average'' means the average across all questions in \datasetname{}.}
    \label{tab:result_en_ar}
\end{table*}

\newpage
\twocolumn
\section{Model Artifacts}
Table~\ref{tab:models} lists the sources of pre-trained models used in this study. All models are sourced from Huggingface \cite{wolf-etal-2020-transformers}.

\begin{table}[h!]
    \centering
    \resizebox{\linewidth}{!}{
        \begin{tabular}{lr}
        \toprule
        \textbf{Models (\#parameters)} & \textbf{Source} \\
        \midrule
        XGLM (564M)  & \texttt{facebook/xglm-564M} \\
        XGLM (1.7B)  & \texttt{facebook/xglm-1.7B}  \\
        XGLM (2.9B)  & \texttt{facebook/xglm-2.9B} \\
        XGLM (4.5B)  & \texttt{facebook/xglm-4.5B} \\
        XGLM (7.5B)  & \texttt{facebook/xglm-7.5B} \\
        \hdashline 
        BLOOMZ (560M)  & \texttt{bigscience/bloomz-560m} \\
        BLOOMZ (1.1B)  & \texttt{bigscience/bloomz-1b1} \\
        BLOOMZ (1.7B)  & \texttt{bigscience/bloomz-1b7} \\
        BLOOMZ (3B)  & \texttt{bigscience/bloomz-3b} \\
        BLOOMZ (7.1B)  & \texttt{bigscience/bloomz-7b1} \\
        \hdashline 
        mT0$_\text{small}$ (300M)   & \texttt{bigscience/mt0-small} \\
        mT0$_\text{base}$ (580M)  & \texttt{bigscience/mt0-base} \\
        mT0$_\text{large}$ (1.2B)   & \texttt{bigscience/mt0-large} \\
        mT0$_\text{xl}$ (3.7B)   & \texttt{bigscience/mt0-xl} \\
        mT0$_\text{xxl}$ (13B)  & \texttt{bigscience/mt0-xxl} \\
        \hdashline 
        LLamA2 (7B)  & \texttt{meta-llama/Llama-2-7b} \\
        LLamA2-chat (7B)  & \texttt{meta-llama/Llama-2-7b-chat} \\
        LLamA2 (13B)  & \texttt{meta-llama/Llama-2-13b} \\
        LLamA2-chat (13B)  & \texttt{meta-llama/Llama-2-13b-chat} \\
        \hdashline 
        Falcon (7B)  & \texttt{tiiuae/falcon-7b} \\
        Falcon-instruct (7B)  & \texttt{tiiuae/falcon-7b-instruct} \\
        Falcon (40B)  & \texttt{tiiuae/falcon-40b} \\
        Falcon-instruct (40B)  & \texttt{tiiuae/falcon-40b-instruct} \\
        \hdashline 
        AraT5 (220M)  & \texttt{UBC-NLP/AraT5-base} \\
        AraT5v2 (220M)  & \texttt{UBC-NLP/AraT5v2-base-1024} \\
        AraGPT2 (1.7BB)  & \texttt{aubmindlab/aragpt2-mega} \\
        \hdashline 
        AceGPT (7B)  & \texttt{FreedomIntelligence/AceGPT-7B} \\
        AceGPT-chat (7B)  & \texttt{FreedomIntelligence/AceGPT-7B-chat} \\
        AceGPT (13B)  & \texttt{FreedomIntelligence/AceGPT-13B} \\
        AceGPT-chat (13B)  & \texttt{FreedomIntelligence/AceGPT-13B-chat} \\
        \hdashline 
        Jais (13B)  & \texttt{core42/jais-13b} \\
        Jais-chat (13B)  & \texttt{core42/jais-13b-chat} \\
        Jais (30B)  & \texttt{core42/jais-30b-v3} \\
        Jais-chat (30B)  & \texttt{core42/jais-30b-chat-v3} \\
        \bottomrule
        \end{tabular}
    }
    \caption{With the exception of GPT-3.5 and GPT-4, all the models used in this study were sourced from Huggingface \cite{wolf-etal-2020-transformers}.}
    \label{tab:models}
\end{table}

\end{document}